%% file: main.tex
\documentclass{article}
\usepackage{iclr_style_preprint,times}

\input{math_commands.tex}

\usepackage{graphicx}
\usepackage{wrapfig}
\usepackage{booktabs}
\usepackage{tabularx}
\usepackage{array}
\usepackage{amsmath,amssymb}
\usepackage{hyperref}
\usepackage{url}
\usepackage{xurl}
\usepackage[most]{tcolorbox}
\usepackage{xcolor}

\title{EvoSim: Learning to Model, Modeling to Learn}

\author{%
  \setlength{\tabcolsep}{0pt}%
  \begin{tabular}{@{}*{3}{>{\centering\arraybackslash\bfseries}p{0.333\textwidth}}@{}}
    Yun-Wei Song\textsuperscript{1,2,3,*} &
    Jin-Kai Tao\textsuperscript{3,4,*} &
    Jun-Dong Zhang\textsuperscript{1,2,*} \\[0.6em]
    Rui Zhang\textsuperscript{5,\dag} &
    Yi-Min Wu\textsuperscript{1,6} &
    Qiang Zhang\textsuperscript{1,2,7,8,\dag} \\[1em]
    \multicolumn{3}{c}{%
      \begin{tabular}{@{}>{\normalfont\small}r@{\hspace{0.45em}}>{\raggedright\arraybackslash\normalfont\small}p{0.82\textwidth}@{}}
        \textsuperscript{1} & Beijing Key Laboratory of Complex Solid-State Batteries, Department of Chemical Engineering, Tsinghua University, Beijing, China \\[0.25em]
        \textsuperscript{2} & State Key Laboratory of Chemical Engineering and Low-Carbon Technology, Tsinghua University, Beijing, China \\[0.25em]
        \textsuperscript{3} & Beijing Tsingyu Technology Co., Ltd., Beijing, China \\[0.25em]
        \textsuperscript{4} & School of Computing and Data Science, The University of Hong Kong, Hong Kong, China \\[0.25em]
        \textsuperscript{5} & Beijing Huairou Laboratory, Beijing, China \\[0.25em]
        \textsuperscript{6} & Tanwei College, Tsinghua University, Beijing, China \\[0.25em]
        \textsuperscript{7} & AI Solid-State Battery Innovation Center, Yibin, China \\[0.25em]
        \textsuperscript{8} & Shanxi Research Institute for Clean Energy, Tsinghua University, Taiyuan, China
      \end{tabular}%
    }
  \end{tabular}%
}

\begin{document}
\maketitle

\begingroup
\renewcommand{\thefootnote}{\fnsymbol{footnote}}
\footnotetext[1]{Equal contribution.}
\footnotetext[2]{Correspondence to: Rui Zhang
  \textless\href{mailto:zhangrui@hrl.ac.cn}{\nolinkurl{zhangrui@hrl.ac.cn}}\textgreater{} and
  Qiang Zhang \textless\href{mailto:zhang-qiang@mails.tsinghua.edu.cn}{\nolinkurl{zhang-qiang@mails.tsinghua.edu.cn}}\textgreater.}
\endgroup

\begin{abstract}
\input{sections/00abstract.tex}
\end{abstract}

\input{sections/01intro.tex}
\input{sections/02related_work.tex}
\input{sections/03method.tex}
\input{sections/04experiment.tex}
\input{sections/05conclusion.tex}
\input{sections/06statements.tex}

\bibliographystyle{iclr_style_preprint}
\bibliography{iclr_style_preprint}

\newpage
\appendix
\input{sections/07appendix.tex}

\end{document}

%% file: math_commands.tex
\usepackage{amsmath,amsfonts,bm}

\def\eqref#1{equation~\ref{#1}}

\def\1{\bm{1}}

\DeclareMathAlphabet{\mathsfit}{\encodingdefault}{\sfdefault}{m}{sl}
\SetMathAlphabet{\mathsfit}{bold}{\encodingdefault}{\sfdefault}{bx}{n}



%% file: sections/00abstract.tex
Physics-based models connect scientific explanation with quantitative prediction. Constructing them requires selecting physical processes, defining states and governing equations, specifying couplings, and identifying parameters from experiments. Existing AI systems remain limited in making these model structure decisions autonomously. We introduce EvoSim, a self-evolving AI scientist for physical modeling. It uses experimental discrepancies to drive mechanism and equation revisions and held-out experimental data to test physical plausibility. Exploration traces make updates to knowledge, skills, and multi-agent orchestration. This co-evolution improves physics-based models and EvoSim’s ability to select mechanisms, diagnose failures, and coordinate research.
We evaluate EvoSim on two industrial battery modeling tasks. It predicts lithium-metal-plating onset from 25 to 45 °C and 2 C to 6 C with a mean absolute error of 1.79\%  in state of charge. Dynamic voltage prediction under vehicle driving conditions achieves a root mean square error of 7.62 mV, surpassing the reported accuracy of models developed by human experts. Self-evolution reduces model and physics errors by approximately 36\% relative to baseline, demonstrating improved scientific modeling capability. EvoSim turns experimental observations into validated models and cumulative research expertise.

%% file: sections/01intro.tex
\section{Introduction}

Physics-based models link physical mechanisms to observable behavior, enabling explanation, prediction, and engineering design. Constructing them requires selecting processes, state variables, equations, couplings, and parameters while preserving consistency, conservation laws, and physical constraints. Unknown mechanisms make this process harder because the assumptions and representations must be chosen appropriately for the model to agree with experimental observations. Physics-based model construction is therefore a long-standing challenge that depends heavily on expert experience.

AI Scientist systems have shown general capabilities for planning and solving scientific problems \citep{Lu2026AIResearchAutomation,Gottweis2026CoScientist}. They can also propose equations in physics-based models and fit them to observations \citep{Shojaee2025LLMSR,Xia2025SRScientist}. Yet these systems usually operate within a model space specified by a human: a candidate mechanism or equation family for symbolic discovery, or a fitting problem with prescribed input and output variables and observations. The agent can therefore apply existing model structures, but cannot autonomously construct new models with explanatory and predictive power.

To improve agents' physical modeling performance, recent systems introduce predefined modeling workflows and procedural knowledge \citep{Ho2026AutoConfiguringSimulators,Li2025CoevolutionBattery,Meng2026FermiLink}. However, the intrinsic complexity of physics-based model construction limits the effectiveness of fixed workflows and preloaded knowledge: the relevant knowledge and sequence of actions must evolve as new evidence changes the modeling problem \citep{Havranek2026MultiAgentDebate,Jwalapuram2026MultiAgentAdvantage}. A more flexible system is therefore needed, one that learns from execution evidence and evolves its modeling strategies.

\begin{wrapfigure}[24]{r}{0.42\textwidth}
\centering
\includegraphics[width=\linewidth]{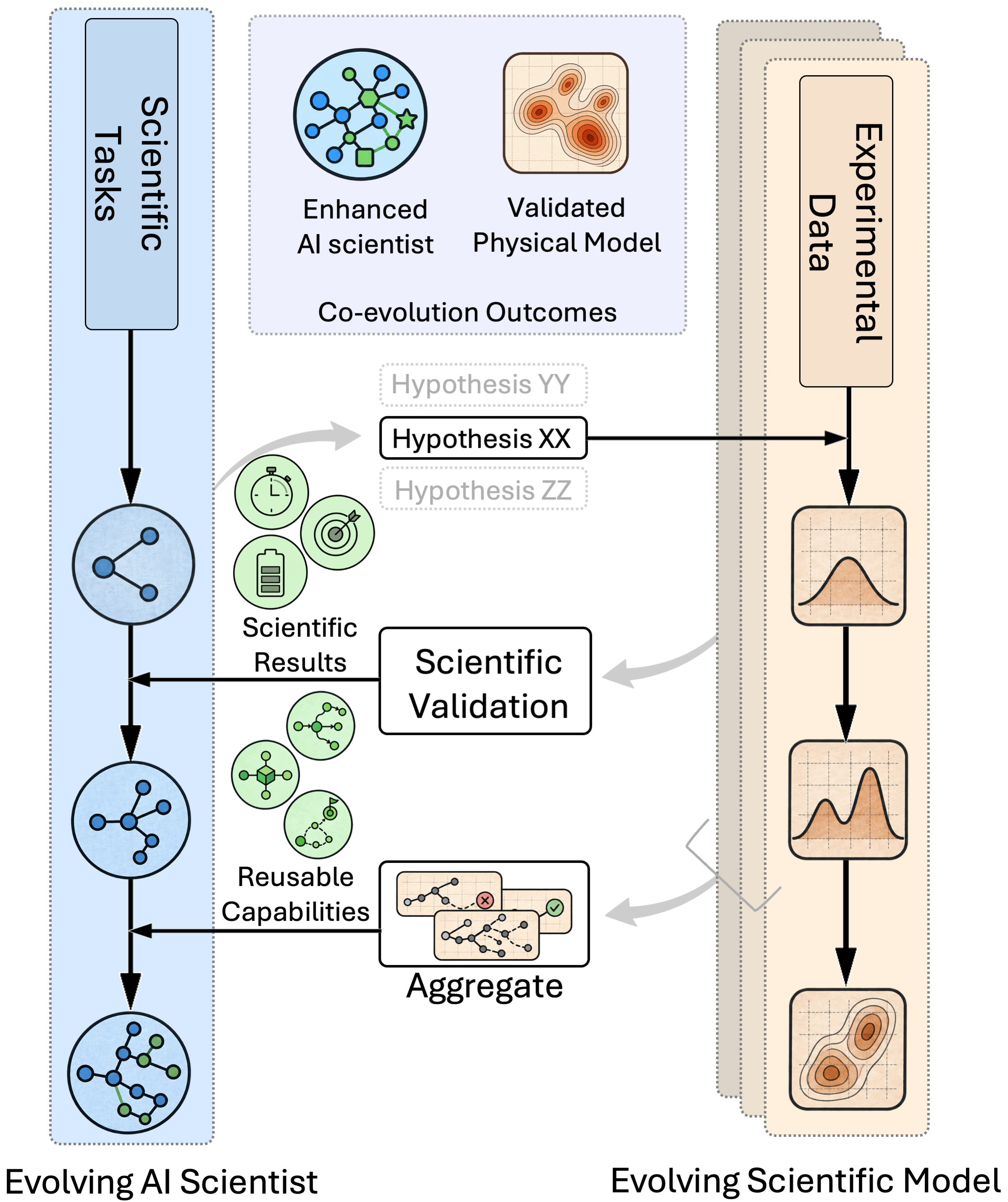}
\caption{EvoSim co-evolves a physics-based model and the AI scientist that develops it.}
\label{fig:coevolution}
\end{wrapfigure}

We herein introduce \textbf{EvoSim}, an AI scientist framework that autonomously constructs and develops physics-based models through this co-evolution process. Starting from a research task and an initial model where available, EvoSim constructs candidate models and uses discrepancies with measurements to guide further hypotheses. It explores alternative mechanisms, states, equations, couplings, and parameters through parallel investigations. Each candidate is checked for correct implementation, successful numerical solution, and agreement with measurements. The results guide model selection and subsequent research. EvoSim also reviews the investigation records to extract domain knowledge, modeling skills, tool procedures, and strategies for coordinating research tasks. These capabilities are assessed for supporting evidence, actionability, and scope of use, separately from decisions about candidate models. Retained capabilities guide subsequent hypothesis selection, model construction, and validation. A rejected model can therefore yield a useful research method, allowing subsequent exploration to build on previous attempts.

We evaluate EvoSim on two industrial battery-modeling tasks. Predicting lithium-plating onset helps define operating limits for fast charging, while predicting dynamic voltage supports assessment of battery behavior under vehicle loads. The plating-onset model achieves a mean absolute error of 1.79\% points in state of charge across held-out validation conditions. The voltage model achieves a root mean squared error of 7.62\,mV on the 30\,$^\circ$C dynamic stress test condition, which is numerically below the compared literature values. Across 45 matched comparisons, capability learning reduces the incidence of model-construction or physical-mechanism failures from 73.3\% to 46.7\%. Together, these results support autonomous modeling in which experimental feedback improves physical explanations, and each investigation builds the research capabilities.

\noindent\textbf{The key contributions are as follows:}
\begin{enumerate}

    \item \textbf{Co-evolution for autonomous physics-based modeling.} EvoSim couples experimental model development with research capability learning. Model investigations identify learning requirements, and retained knowledge and methods support further exploration of physical mechanisms.
    \item \textbf{Parallel exploration of physical mechanisms and model structures.} EvoSim investigates competing explanations for prediction discrepancies in parallel by changing mechanisms, equations, and couplings. Each candidate is checked for correct implementation, solved, and compared with measurements. These results guide subsequent exploration, allowing the system to investigate a broad range of physics-based models and progressively refine them through experimental feedback.
    \item \textbf{Learning from model investigations to improve subsequent modeling.} EvoSim extracts knowledge and procedures from successful and failed attempts, checks their supporting evidence and scope of use, and applies them to subsequent model construction and validation. On a voltage-modeling task, this learning mechanism produces a validated model with 22.015\,mV RMSE within three submissions, reducing error by 75.4\% compared with providing a chronological summary of previous attempts. That baseline does not reach comparable accuracy even after twelve submissions.
    \item \textbf{Predictive accuracy and improved research reliability in industrial tasks.} Two battery tasks demonstrate model accuracy. Across 45 matched comparisons, capability learning reduces model-construction or physical-mechanism failures by 36.4\% relative to the raw condition and reduces active runtime to 57.4\% of that condition.
\end{enumerate}

%% file: sections/02related_work.tex
\section{Related Work}

\subsection{Autonomous Scientific Research}

Autonomous scientific research involves deciding what to investigate as well as executing experiments. AI-scientist systems coordinate hypothesis generation, experimentation, and refinement \citep{Lu2026AIResearchAutomation,Swanson2025VirtualLab,Gottweis2026CoScientist}. Related modeling work studies equation discovery \citep{Shojaee2025LLMSR,Xia2025SRScientist,Zheng2025NewtonBench} and simulation-based problem solving \citep{Mudur2025FEABench,Meng2026FermiLink}. Li et al., for example, connect language models to prebuilt battery models \citep{Li2025CoevolutionBattery}. However, executing a simulation does not resolve which physical assumptions should change when an existing model disagrees with measurements. In contrast to using an existing formulation, EvoSim makes the formulation itself subject to revision. It tests competing physical explanations through changes to equations, states, and couplings, and uses the resulting evidence to revise the investigation plan. Its autonomy thus concerns both what model to build and what investigation to pursue, rather than simulation execution alone.
\subsection{Scientific Evolution and Self-Improvement}

Self-improving agents turn execution feedback into changes to their code, context, or research strategies. DGM evolves agent code, while ACE accumulates and revises contextual strategies \citep{Zhang2026DGM,Zhang2026ACE}. EvoScientist extends experience reuse to scientific ideation and experimentation \citep{Lyu2026EvoScientist}, and SIGA optimizes simulator adapters offline from logged trajectories \citep{Ho2026AutoConfiguringSimulators}. EvoSim couples such capability updates to the continuing development of a physics-based model. Compared with distilling research histories into reusable strategies, Scientific Evolution explicitly distinguishes observed failures, proposed corrections, and procedures supported by application. A correction is retained as a procedure only when its application supports the expected effect under the recorded conditions. Model acceptance is decided separately, allowing a rejected candidate to improve research methods without endorsing its physical explanation. These methods guide subsequent hypotheses, model construction, and investigation plans. Their outcomes then inform further updates to both the physics-based model and the scientist, coupling model evolution with recursive self-improvement.

%% file: sections/03method.tex
\section{Method}

\subsection{Architecture}

\begin{figure}[t]
\centering
\includegraphics[width=\linewidth]{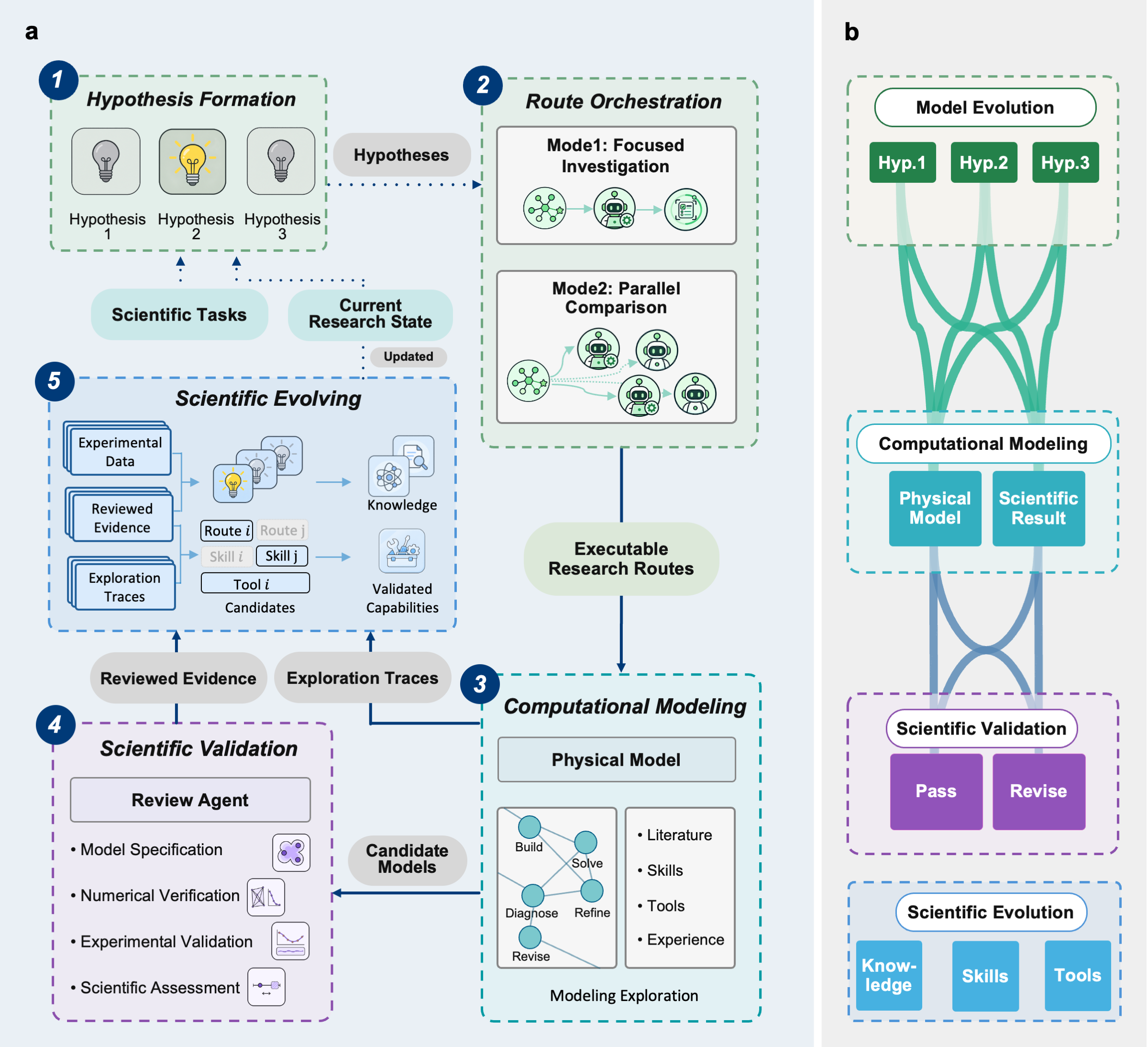}
\caption{\textbf{EvoSim's architecture and coupled evolution.}
\textbf{(a)} The research loop connects hypothesis formation, research planning, model construction, validation, and Scientific Evolution.
\textbf{(b)} Model evolution develops and evaluates alternative hypotheses through computational modeling and scientific validation. Scientific evolution uses feedback from past investigations to update knowledge, skills, tools, and research strategies.}
\label{fig:architecture}
\end{figure}

EvoSim treats scientific modeling as the coupled evolution of a physics-based model and the AI scientist's research capability. Each investigation informs two distinct decisions: how to develop the model and what to retain or revise in the scientist's knowledge and research methods. A candidate that is not adopted can still reveal a failure mode or motivate changes to the procedures and strategies used in later investigations. The updated scientist then helps produce the evidence for further model development and for its own next update.

Figure~\ref{fig:architecture}a shows the architecture of EvoSim. Hypothesis Formation proposes promising explanations, and Route Orchestration organizes focused investigations and parallel comparisons. Computational Modeling builds and runs candidate models. Scientific Validation examines their implementation, solutions, and agreement with experimental measurements. Scientific Evolution leverages reviewed evidence and exploratory trajectories to drive the AI scientists' capability evolution.

Each task specifies a scientific question, a target quantity, experimental measurements and a solver. Research proceeds in rounds indexed by $t$. Each round groups the investigations organized under one research plan.

At round $t$, the current model is
\begin{equation}
M_t=(P_t,\theta_t),
\label{eq:model_state}
\end{equation}
where $P_t$ specifies the executable physics-based formulation and $\theta_t$ contains its parameter values. The formulation defines the physical processes, state variables, equations, couplings, forms of the initial and boundary conditions.

Let $\hat{y}_t$ denote the model predictions and $y^{\mathrm{exp}}$ the corresponding experimental measurements. The residual is $\lambda_t=\hat{y}_t-y^{\mathrm{exp}}$.

The research state $R_t$ is the global record of the current task. It links hypotheses to candidate branches and records completed investigations, reviewed evidence, and open questions. EvoSim uses this record to coordinate work and build on findings from other branches.

The scientist state $S_t$ contains physics-based findings, failure experience, and research methods retained for later use. These include modeling skills, tools and tool-use procedures, validation checks, and strategies for organizing investigations. Scientific Evolution reviews the investigations recorded in $R_t$ to derive findings and methods with conditions of use. Thus, $R_t$ records what was investigated and observed, while $S_t$ guides how subsequent research is conducted.

An investigation can leave $M_{t+1}=M_t$ while producing $S_{t+1}\neq S_t$.

\subsection{Model Evolution}

\textit{Hypothesis Formation} uses the residual $\lambda_t$, physics-based findings in $S_t$, and earlier investigations in $R_t$ to propose explanations for the difference between model predictions and experimental measurements. Each hypothesis specifies a model change and its expected effect on the predictions. One explanation may require coordinated changes to equations, states, couplings, or parameter values. Distinct explanations form separate candidate branches.

\textit{Route Orchestration} turns these hypotheses and open questions into a research plan. Given the current state of research $R_t$, AI scientists can evolve physics-based models in terms of both depth and breadth. Focused Investigation addresses a specific research need, such as task interpretation, theoretical analysis, exploratory modeling, or failure diagnosis. Its findings inform candidate hypotheses and their evaluation. Parallel Comparison explores different explanations or modeling approaches concurrently and evaluates them against the same experimental measurements under common criteria. Within the hierarchical agent architecture, the plan arranges dependent investigations sequentially and independent alternatives in parallel. It specifies the proposed changes, conditions held fixed, evaluation procedures, and computational budget.

Let $h_{t,j}$ denote the hypothesis for candidate $j$, let $m_t$ be the number of model candidates, and let $\rho_t$ denote the research plan. The operation $\pi_{S_t}$ represents the hypothesis and planning decisions made using the current scientist state:
\begin{equation}
\begin{aligned}
\left(\{h_{t,j}\}_{j=1}^{m_t},\rho_t\right)
&= \pi_{S_t}(M_t,\lambda_t,R_t),\\
\left(M'_{t,j},\tau_{t,j}\right)
&= \mathcal{I}(M_t,h_{t,j},\rho_t;S_t).
\end{aligned}
\label{eq:research_round}
\end{equation}
Here, $\mathcal{I}$ denotes \textit{Computational Modeling}. It implements each candidate on a copy of $M_t$, runs the solver, and diagnoses implementation or numerical problems as they arise. The exploration trace $\tau_{t,j}$ records actions, intermediate results, and changes made during this process. It connects the proposed hypothesis to the implemented model and the results obtained from it.

\textit{Scientific Validation} first checks the executed model against the proposed specification, including whether the intended changes are active. Numerical verification examines solver status and the required outputs. For numerically valid runs, experimental validation compares model predictions with experimental measurements using the task's error metric. Scientific assessment examines what the results support about the proposed explanation and its tested conditions.

Let $E_{t,j}$ denote the reviewed evidence from these checks and assessments. Set $a_{t,j}=1$ when the candidate meets the task's acceptance criteria specified in $\rho_t$, and $a_{t,j}=0$ otherwise. The model decision is
\begin{equation}
M_{t+1}
=
\operatorname{Select}_{\rho_t}\!\left(
M_t,
\{(M'_{t,j},E_{t,j}):a_{t,j}=1\}
\right).
\label{eq:model_update}
\end{equation}
The operation $\operatorname{Select}_{\rho_t}$ applies the task's comparison rule to determine the model used for subsequent research. If no candidate meets the acceptance criteria, EvoSim retains $M_t$.

The results of all investigations update $R_{t+1}$, including those whose candidate models are not adopted. Later rounds use this shared record to revise explanations, investigate unresolved failures, and develop new candidates from $M_{t+1}$. Findings from separate branches can support a combined hypothesis, which is implemented and evaluated as a new candidate. Parallel branches thus contribute both candidate models and evidence that guides further exploration.

\subsection{Scientific Evolution}

\textit{Scientific Evolution} updates the scientist's domain knowledge, procedures for modeling and validation, and strategies for organizing investigations. It uses reviewed evidence and exploration traces to derive findings, experience, and research methods with conditions of use. These updates draw on both candidate evaluations and past investigations.

For a failed attempt, EvoSim identifies where the observed result departed from the expected outcome and records the attempted change, the failure, and its conditions. These problems guide diagnosis of the modeling procedure. Discrepancies in numerically valid predictions motivate further tests of the physical explanation. When the evidence supports a cause, that explanation is retained with the failure observation. The observed failure and its conditions can guide later diagnosis even while its cause remains unresolved.

EvoSim uses this analysis to propose a correction to a modeling step or research decision. The correction specifies what should change and the expected effect to check when it is applied. An untested correction remains a proposal for further investigation. When its application provides evidence supporting the expected effect, Scientific Evolution retains it as a reusable procedure with its conditions of use. Otherwise, the new feedback guides further diagnosis and revision.

All investigations contribute physics-based findings, procedures, and validation checks tied to the conditions under which they were supported. EvoSim retains these findings and methods rather than treating an entire successful modeling route as a procedure to repeat unchanged. Both successful and failed investigations therefore inform how the scientist approaches subsequent research.

The research strategies retained in $S_t$ guide how investigations are organized within the flexible agent hierarchy. They inform the construction of the round-specific plan $\rho_t$. Feedback from earlier work guides revisions to when focused analysis is needed, which alternatives to compare in parallel, and which checks should precede further modeling. The revised strategies are applied in subsequent investigations and reassessed using their outcomes.

Let $\Delta S_t^{+}$ contain the physics-based findings, failure experience, and new or revised methods retained from the round on the basis of this review. Let $\Delta S_t^{-}$ contain earlier entries replaced by these revisions. The scientist update is
\begin{equation}
S_{t+1}
=
(S_t\setminus\Delta S_t^{-})\cup\Delta S_t^{+}.
\label{eq:scientist_update}
\end{equation}
Replacing a procedure does not require removing the failure experience that motivated its revision: later research can use both the conditions associated with the failure and a tested way to address it.

At the next round, EvoSim retrieves knowledge and methods whose conditions match the current research problem. Physics-based findings inform the explanations to test, failure experience guides what to examine, and research strategies guide how investigations are organized:
\begin{equation}
\left(\{h_{t+1,j}\}_{j=1}^{m_{t+1}},\rho_{t+1}\right)
=
\pi_{S_{t+1}}(M_{t+1},\lambda_{t+1},R_{t+1}).
\label{eq:next_round}
\end{equation}
The retrieved methods also guide the implementation and validation of the next candidate models.

These investigations produce new feedback about both the developing model and the methods used to study it. Scientific Evolution uses this feedback to revise methods, refine their conditions of use, and investigate newly observed failures. The updated scientist thus helps conduct the research that produces evidence for its own further improvement, coupling model evolution with recursive self-improvement of research capability.

%% file: sections/04experiment.tex
\section{Experiments}

\subsection{Experimental Setup}

\textbf{Evaluation.} We study two industrial battery-modeling tasks. We examine the resulting physics-based models and their predictions. Comparisons then test whether Scientific Evolution improves later investigations. Repeated-task comparisons measure failures and execution time. We also examine how evaluation results affect the scientific findings selected by EvoSim.

\textbf{Execution.} EvoSim uses COMSOL Multiphysics\textsuperscript{\textregistered} to construct and solve the physics-based models. Each model undergoes three checks. The implementation check confirms the active equations and states. The solution check confirms that the requested simulation is complete and the outputs are finite. The measurement check compares the prediction with the specified observations. Appendix~\ref{sec:appendix-measurement-role} gives the evaluation procedure.

\subsection{Industrial Battery Modeling}

\begin{figure}[t]
\centering
\includegraphics[width=\linewidth]{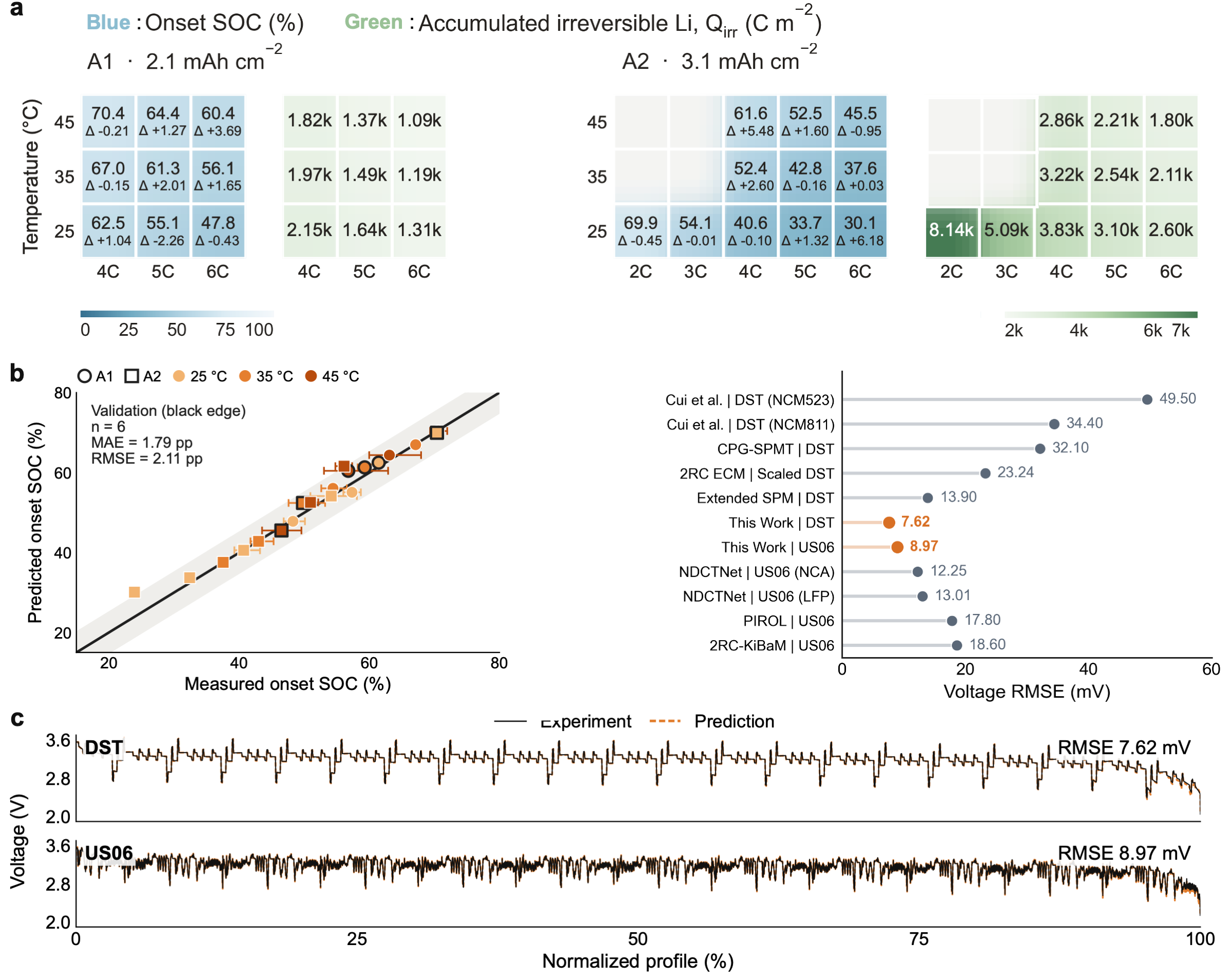}
\caption{\textbf{Lithium-plating onset and dynamic-voltage predictions.} \textbf{(a)} Onset SOC and accumulated irreversible lithium for two graphite-electrode designs across temperatures and charging rates. \textbf{(b)} Measured and predicted onset SOC for 20 conditions (left), and voltage RMSE compared with reference values (right). The voltage panel reports cross-validation errors for 30$^\circ$C DST (7.62\,mV) and 20$^\circ$C US06 (8.97\,mV). \textbf{(c)} Complete measured and predicted voltage traces for the same two cross-validation results.}
\label{fig:results}
\end{figure}

\paragraph{1. Lithium-plating onset during fast charging.} Fast charging reduces waiting time. At high charging rates, lithium can deposit as metal on the graphite anode surface rather than enter the graphite. This process is called lithium plating. The deposits can form harmful tree-like structures. Predicting plating onset helps determine charging rates that do not compromise battery life or safety.

\textbf{Data.} We use 20 onset measurements from 61 Li$\mid$graphite cells \citep{Konz2023HighThroughputLiPlating}. These cells pair a lithium-metal electrode with a graphite electrode. The measurements cover two graphite-electrode designs, A1 and A2, at temperatures of 25, 35, and 45$^\circ$C. Charging rates range from 2C to 6C. C-rate expresses charging current relative to rated cell capacity. The experimental target is the graphite state of charge (SOC) at which accumulated irreversible lithium reaches 0.05\% of graphite capacity. This definition connects electrode design and charging conditions to fast-charge limits.

\textbf{Model.} EvoSim starts from a one-dimensional porous-electrode model for electrolyte transport, electronic conduction, particle diffusion, and graphite intercalation. It adds a parallel plating and stripping reaction driven by the local plating overpotential. Onset is defined by a capacity-normalized accumulated irreversible-lithium threshold. The 14 training conditions identify a shared plating exchange-current scale and a bounded rate-dependent fraction that converts accumulated deposition into irreversible lithium. The model is then fixed before prediction of the six validation conditions.

\textbf{Results.} The six validation conditions are excluded from parameter identification. On these held-out conditions, predicted onset SOC has an MAE of 1.79 SOC percentage points and an RMSE of 2.11 SOC percentage points (Figure~\ref{fig:results}a,b). Appendix~\ref{sec:appendix-representative-models} provides the equations, onset criterion, development path, solution checks, and pointwise results.

\paragraph{2. Dynamic voltage under vehicle driving conditions.} During driving, a battery supplies power for acceleration and receives energy recovered during braking. These changes in current cause its voltage to vary. Predicting these voltage changes helps determine whether the battery can meet driving demands while remaining within its operating limits.

\textbf{Data.} We use the A123 A1-007 data for Dynamic Stress Test (DST) and US06 profiles\citep{CALCEA123BatteryData,Guo2026CPGSPMTDataset}. The data contain current, measured surface temperature, and terminal voltage. The traces contain 109,746 time-aligned samples recorded at one-second intervals. They cover 30.48 hours of operation. The traces include rapid load changes, current reversals, rest periods, and voltage recovery. The physics-based model must describe the voltage response throughout each complete driving profile.

\textbf{Model.} EvoSim starts with four electrode filling states and a voltage equation for equilibrium potentials, charge-transfer polarization, and internal resistance. It rebuilds the states with the exact zero-order-hold update, then adds electrolyte transport and interfacial charging to the solved physics-based model. Terminal voltage follows from the solved electrochemical potentials and resistive losses. Appendix~\ref{sec:appendix-dynamic-voltage} gives the model and held-out evaluation.

\textbf{Results.} We evaluate the fixed physics-based model by leaving out one temperature at a time. For each held-out temperature, the model uses the other temperatures for parameter identification. Figure~\ref{fig:results}c shows two complete predicted and measured voltage traces at held-out temperatures.

\begin{figure}[t]
\centering
\includegraphics[width=\linewidth]{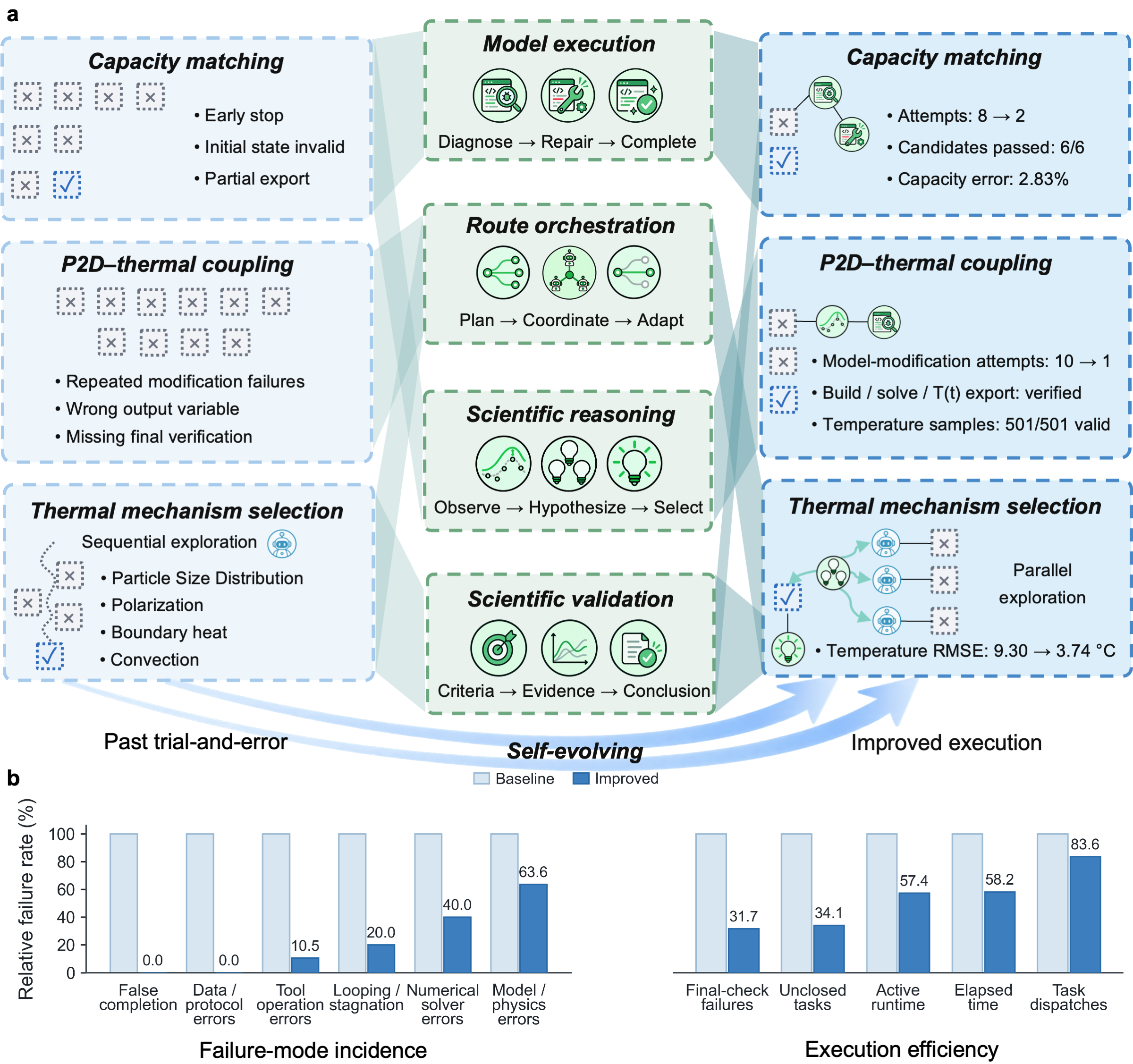}
\caption{\textbf{Effect of Scientific Evolution.} \textbf{(a)} Baseline investigations and investigations using saved procedures on three tasks. \textbf{(b)} Failure counts and execution measures across 45 matched pairs from 11 modeling tasks. Each measure is normalized to its corresponding Baseline value, shown as 100\%. The figure label Improved denotes EvoSim with Scientific Evolution.}
\label{fig:evolution}
\end{figure}

\subsection{Effect of Scientific Evolution}

\textbf{Controlled comparison.} We use the 30$^\circ$C DST trace, containing 7,461 samples, to compare three forms of prior information. All three conditions use the same task, starting model, data, solver, tools, checks, and a maximum of 12 attempts. The main comparison uses the first three attempts. Later attempts test whether continued search changes the result. \emph{Baseline} receives no earlier record. \emph{Naive Memory} receives a chronological summary of the earlier attempts and results. \emph{Scientific Evolution} receives a checked modeling procedure from the same investigation. The accepted model must implement the required states and terminal-voltage equation. It must also export a complete finite time series. Scientific Evolution reaches its best accepted model on attempt 2. Its experimental-voltage RMSE is 75.4\% lower than the Baseline value. Naive Memory remains close to Baseline (Table~\ref{tab:scientific_learning_ablation}). Appendix~\ref{sec:appendix-memory-ablation} reports each model change and its checks. Appendix~\ref{sec:appendix-capability-trace} shows how the procedure was obtained and used later.

\begin{table}[t]
\centering
\caption{\textbf{Ablation on the four-state voltage task.} Performance and progress during the model search.}
\label{tab:scientific_learning_ablation}
\begin{tabularx}{\linewidth}{@{}l>{\centering\arraybackslash}p{0.25\linewidth}>{\raggedright\arraybackslash}X@{}}
\toprule
Condition & Best valid RMSE (mV) & Progress during search \\
\midrule
Baseline & 89.478 & Unchanged through 12 attempts \\
Naive Memory & 89.383 & Unchanged through 12 attempts \\
Scientific Evolution & \textbf{22.015} & Attained on attempt 2 \\
\bottomrule
\end{tabularx}
\end{table}

\textbf{Later investigations.} Three records show how a procedure from an earlier investigation changes work on a related task (Figure~\ref{fig:evolution}a and Table~\ref{tab:appendix-cross-task-reuse}). Each later task has its own instruction, starting model, data, simulator, tools, and checks. In the capacity-matching task, the saved matching and stopping steps reduce the required attempts from 8 to 2. A procedure from the thermal-mechanism investigation is later used in the P2D--thermal investigation. It compares independent physical mechanisms before further model changes. The number of model changes decreases from 10 to 1. In the thermal-mechanism task, the same comparison procedure accompanies a decrease in temperature RMSE from 9.30 to 3.74$^\circ$C. These records connect an earlier result to a procedure update, the next modeling decision, and a measured later outcome. Appendix~\ref{sec:appendix-four-comparisons}, ~\ref{sec:appendix-capability-atlas} give the complete records.

\textbf{Reliability and efficiency.} We analyze 45 matched pairs across 11 modeling tasks. Each pair contains one Baseline run and one Scientific Evolution run. The two runs use the same task, starting model, data, solver, checks, and evaluation budget. Model and physical-mechanism failures decrease from 33 to 21 runs. Among pairs with different outcomes, 15 favor Scientific Evolution and 3 favor Baseline. The two-sided exact McNemar test gives $p=0.0075$. Adjustment across the six failure modes gives $p=0.0151$. Active execution time for these later modeling runs decreases by 42.6\% (Figure~\ref{fig:evolution}b). Appendix~\ref{sec:appendix-paired-analysis} gives the failure definitions, statistical test, counts, and execution totals.

\textbf{Scientific conclusions.} Across nine evaluated tasks, four conclusions apply to the tested conditions. Four conclusions apply only to a narrower range. One proposed general mechanism is rejected because every tested model fails the evaluation criterion. All nine decisions agree with the evaluation results. Appendix~\ref{sec:appendix-cross-task-validation} gives the selection criteria and results.

%% file: sections/05conclusion.tex
\section{Conclusion}

We presented EvoSim, an AI scientist that co-evolves physics-based models and reusable modeling capabilities. The model evolving process investigates physical hypotheses in parallel, using experimental discrepancies to guide changes to mechanisms, equations, and couplings. The scientific evolving process extracts knowledge, procedures, and research strategies from successful and failed attempts, checks their supporting evidence, and applies them to subsequent investigations.

Two industrial tasks demonstrate the effectiveness of this approach. The battery plating-onset model achieves a low MAE of 1.79\% in state of charge across experimental measured conditions. The battery dynamic-voltage model reaches 7.62\,mV RMSE. Across 45 matched comparisons, it reduces model-construction or physics-based mechanism failures by 36.4\%. These results support improvements in both model accuracy and modeling efficiency and reliability, and offers a path toward autonomous scientific intelligence to understand the physics world.

%% file: sections/06statements.tex
\section*{Acknowledgments}

This work was supported by the National Natural Science Foundation of China (Grant No. 223B1012) and the Tsinghua University Initiative Scientific Research Program.

%% file: sections/07appendix.tex
\section{Modeling Setup and Evaluation}
\label{sec:appendix-rsi-environment}

Each experiment uses a controlled modeling setup. The battery cell, protocol, observations, starting equations, permitted model modifications, COMSOL solver, output quantities, comparison metric, and stopping rule are fixed before the agent begins. A candidate may contain one or more permitted modifications, depending on the task. Candidates can be explored through a serial continuation, a parallel set of alternatives, or both. The solver returns the resulting fields, states, voltage, or temperature. A research record links each candidate's modifications to the evidence and decision. A run is one complete execution of a task. FEABench also combines a task specification, a multiphysics solver, execution feedback, and an evaluator. SIGA separates the supplied information, tools, stored procedures, and stopping checks \citep{Mudur2025FEABench,Ho2026AutoConfiguringSimulators}. Another battery study connects a language agent to a physics-based model \citep{Li2025CoevolutionBattery}.

\subsection{Controlled Modeling Problem}

Each run has a written task, an executable starting model, observation data, a simulator, permitted model changes, a work budget, and fixed acceptance rules. Some tasks also contain separate data for the final comparison. These items determine what the agent can read, what it can change, and what result the model must produce. A candidate model is one modified copy of the starting or currently selected model.

The fixed task inputs contain the task description, starting-model source, data, required outputs, checks, and budget. These files are read-only during evaluation. A research record is a saved account of a candidate modification or modification set. It contains the reason for the modification, the result, and the decision to keep or reject it.

Table~\ref{tab:appendix-rsi-environment} lists the inputs and records used in the reported runs. All stated observations are available for model construction. When a task includes a separate comparison target, its role is fixed before a candidate is selected.

\subsection{Starting Physics-Based Model and Fixed Inputs}

The physical experiment supplies the cell, measurement protocol, instruments, and voltage or temperature records. These observations are the fixed evidence used during model development. The modeling setup supplies an executable representation of the cell and a controlled way to change it. For a battery task, the fixed inputs specify the physics-based model or Java model-building script, geometry, physics interfaces, parameters, initial and boundary conditions, input time range, target quantities, allowed edits, output format, and stopping rule. They also specify any reference implementation used for diagnosis and any comparison target used for evaluation.

The multiphysics solver supplies the model states and output files. A reference implementation can be used for diagnosis when the task allows it. Section~\ref{sec:appendix-dynamic-voltage} records each added dynamic state and its connection to the electrochemical quantities. The measurements remain the reference for constructing and evaluating the model.

\begin{table}[t]
\centering
\caption{Fixed inputs, permitted actions, and saved evidence in the reported experiments.}
\label{tab:appendix-rsi-environment}
\scriptsize
\setlength{\tabcolsep}{3pt}
\renewcommand{\arraystretch}{1.08}
\begin{tabularx}{\linewidth}{@{}>{\raggedright\arraybackslash}p{0.17\linewidth}>{\raggedright\arraybackslash}p{0.29\linewidth}>{\raggedright\arraybackslash}p{0.24\linewidth}>{\raggedright\arraybackslash}X@{}}
\toprule
Part & Fixed before a run & Operation available to the agent & Evidence saved after the run \\
\midrule
Fixed task inputs & Scientific target, starting model, input conditions, required outputs, observations, and run rules & Read the fixed files and propose permitted model modifications & Input-file list, file hashes, and task directory \\
Starting model & Physics-based equations, geometry, parameters, initial conditions, input time range, and relation to the observations & Copy the current model and edit permitted model or solver elements & Submitted Java source or physics-based model file, parameter values, and model-state checks \\
Multiphysics solver & Local or remote solver and the required output format & Build each candidate, solve it, and request the specified fields and states & Solver status, logs, output files, and resource use \\
Research record & Current model, permitted earlier records, saved procedures, and acceptance rule & Interpret feedback and select the next candidate under the same rules & Hypothesis, change, diagnosis, decision, and link to the parent model \\
Checks and metric & Implementation checks, solution checks, measurement metric, and any specified comparison target & Receive the observations and feedback released by the task & Rebuild, solve, export, compare, and record the decision \\
Budget and isolation & Work slots, time and job limits, fixed inputs, and permitted directories & Spend a slot and write only inside the candidate directory & Input-file list, access log, budget record, and saved outputs \\
\bottomrule
\end{tabularx}
\end{table}

\subsection{Modeling Agent and Solver Evaluation}

The modeling agent receives the fixed inputs and permitted earlier records. It can copy the current model, edit permitted equations or settings, build and solve the model, and read the solver fields, states, logs, and output files. The evaluation checks whether the model was rebuilt and solved. It also checks whether the required states and inputs are active, whether the outputs are complete and finite, and how the prediction compares with the observations. The agent then records its hypothesis, diagnosis, and next model decision.

The same interface is used in all comparison conditions. Baseline receives no earlier research record. Naive Memory receives a chronological summary of an earlier investigation. Scientific Evolution receives a procedure that EvoSim produced and checked from that investigation. The comparison tests whether converting the earlier evidence into a specific operation and checks changes the next model decision. The solver, input trace, starting model, output format, checks, and comparison point remain fixed.

\paragraph{Language model and prompts.}
The controlled three-condition comparison uses \texttt{gpt-5.5} in all conditions. The model version, system prompt, task prompt, tools, submission budget, and stopping rule are fixed across the three runs. The planning prompt asks the agent to inspect the task, current model, and available evidence; select the next candidate modification or modification set; state its expected physical or numerical effect; and define the checks needed after execution. The execution prompt asks the agent to implement the selected candidate, run the solver, export the required quantities, and report the model files and results. The final-review prompt starts in a separate session. It reads the candidate model, source files, solver log, exported results, and fixed acceptance rules. It returns \textsc{pass} only when the implementation, solution, and measurement checks all pass.

\paragraph{Procedure retrieval and update.}
Each saved procedure states the model type, required inputs, supported conditions, action, and checks. The action can be a model change, calculation, or comparison. Before choosing the next model change, EvoSim retrieves a procedure only when its conditions match the current task. A later result can extend the supported conditions, add a failed condition, or replace an earlier action. The new record keeps the model version, result files, and review decision that justify the change. If the new evidence does not support a revision, the existing procedure remains unchanged.

\subsection{Isolation and Saved Model Records}

Every candidate is evaluated in a separate directory. The directory contains the fixed task inputs, model files, Java support files, results, logs, references, and research record. The system lists the fixed inputs before execution and checks them again afterward. Reads and writes are limited to the permitted directories. Local and remote solver runs use the same file layout. The solver job record, output files, and log are saved with the candidate.

Each candidate is stored with its execution files. A selected candidate is linked to its parent model, code changes, parameter values, solver log, output files, comparison table, error metric, diagnosis, and decision. A rejected candidate remains in the research record and cannot replace the selected model. A run receives earlier research records only when the experiment allows them. Solver sessions, other candidates, and unrecorded information remain unavailable.

\subsection{Evaluation Checks, Budget, and Termination}

Candidate acceptance uses three checks in sequence: implementation, solution, and agreement with measurements. Together, these checks form Scientific Validation. The implementation check confirms that the submitted Java source or physics-based model can be rebuilt. It also confirms that the required physics and state variables enter the active equations. Inputs, initial conditions, time range, and output expressions must be connected correctly. The solution check requires completion of the requested time range or of an event defined before execution. It also requires finite values and valid output files. The measurement check applies the fixed metric to the specified observations. A candidate replaces the current model after all required checks pass. A failed candidate remains available for diagnosis.

The budget is part of the environment. Each controlled multiphysics-solver submission uses one slot in the run budget. This includes a submission that fails during checking, compilation, solving, or export. Local compilation that is not submitted remains outside the solver budget. Candidates are submitted one at a time. A run stops according to its stopping rule or when the budget is exhausted. If the budget ends first, the run reports the best result permitted by that task's rules. Each comparison condition uses its stated budget, solver access, and time limits.

\section{System and Research Procedure}

The rest of the appendix has four parts. The lithium-plating case shows how EvoSim adds a missing physical process and evaluates it on measured conditions. The dynamic-voltage case defines a solved electrochemical model and evaluates its complete voltage traces. The Scientific Evolution comparison shows how a checked procedure changes a later model decision. The final section links fixed models to conclusions under stated conditions.

The first subsection gives the research procedure. The second explains how Scientific Evolution records a checked modeling procedure and uses it again.

\subsection{Research Procedure}

Each exploration step creates one or more candidate copies of the current model and a research record for each candidate. Before execution, each record states the parent model, model changes, expected effect, observations, fixed quantities, required outputs, metric, and acceptance rule. Candidate copies may run serially or in parallel, with each candidate in its own modeling session. After execution, the record adds the implemented changes, commands, build and solve status, solver states, output files, numerical result, model decision, and decision to save or reject the procedure. A later step can start from a saved model, an explicitly selected branch, or another candidate defined by the task. Every candidate remains linked to its files, result, and diagnosis.

\subsection{Reusable Procedures and Later Use}

Scientific Evolution saves a modeling procedure after it passes a stated check. The record gives the earlier result, action, expected response, checks, and supported conditions. When EvoSim uses the procedure again, it records the later model decision and result. Table~\ref{tab:appendix-capability-trace} follows one procedure from its first result to a later model update.

\section{Battery Modeling Cases}
\label{sec:appendix-representative-models}

The two cases require different kinds of model development. Lithium-plating onset defines a fast-charge boundary that affects safety, lifetime, and electrode design. Its starting point is a basic Doyle--Fuller--Newman/pseudo-two-dimensional (DFN/P2D) model with electrolyte transport, electronic conduction, particle diffusion, and intercalation. The appendix specifies the 20-condition onset matrix, the accumulated irreversible-lithium threshold, the added plating/stripping reaction, and the retained deposition state. Dynamic-voltage modeling supports state estimation, power prediction, and control in battery-management systems. It requires electrochemical states to remain valid through long vehicle-load traces and changing temperatures. Both cases provide public measurements, executable starting equations, and quantitative outputs. The plating case reports agreement with the measured matrix. The voltage case reports a solved reduced electrochemical model and its complete-trace run results.

\subsection{Role of Measurements in Physics-Based Model Development}
\label{sec:appendix-measurement-role}

The object being evaluated is a physics-based model: its state variables, governing equations, physical couplings, identified parameters, and solved outputs. Experimental measurements constrain this model. They show which physical behavior the current equations fail to describe, provide values for parameter identification, and supply quantitative comparisons after the model is solved. The selected model must also pass implementation, solution, conservation, state-limit, and complete-output checks.

The plating-onset matrix contains 20 measured conditions. A fixed split assigns 14 conditions to training and six to validation. The same split is used throughout model development. EvoSim introduces a rate-dependent mapping from accumulated deposited lithium to irreversible lithium. Its parameters are identified only on the 14 training conditions and are fixed before prediction of the six validation conditions.

The dynamic-voltage case evaluates one fixed model structure by leave-one-temperature cross-validation on complete driving traces. The evaluation contains eight temperature folds and 16 traces. In each fold, both the DST and US06 traces at the held-out temperature are excluded from parameter identification, and both are evaluated after the fold-specific parameters are identified. We report two pre-defined individual-condition results in the main text. Section~\ref{sec:appendix-dynamic-voltage} gives the split definition and the reported held-out results.

These checks answer different scientific questions. Agreement with the development measurements tests whether the constructed equations and calibrated parameters describe the observed cell. Solution and physical checks test whether the implemented model solves the stated equations and respects their constraints. Cross-validation on separate curve groups tests calibration of the fixed model structure. The paper reports each result according to this role.

\subsection{Shared Model Development Sequence}

The shared development sequence is described in the case sections below. In the plating case, the agent constructs the DFN/P2D foundation, checks transport and intercalation, introduces a missing electrochemical reaction, and connects deposited lithium to the measured onset definition. In the voltage case, the agent establishes the four-state physical baseline, then extends the solved equations with electrolyte transport and interfacial charging.

\paragraph{Plating sequence.} The plating case starts from a basic DFN/P2D description of a Li$\mid$graphite half-cell. EvoSim specifies geometry, electrolyte transport, solid diffusion, graphite equilibrium potential, and intercalation kinetics for electrode designs A1 and A2. It then adds a parallel plating/stripping reaction and an accumulated irreversible-lithium state. A rate-dependent mapping is identified on 14 training conditions and evaluated on the six fixed validation conditions.

\paragraph{Voltage sequence.} The voltage case starts from four electrode filling states and a voltage equation containing equilibrium, charge-transfer, and ohmic terms. EvoSim rebuilds the four states with the exact discrete update and establishes the physical baseline. It then extends the solved equations with electrolyte transport and interfacial charging. The terminal voltage is obtained from the solved electrochemical potentials and resistive losses. The complete held-out evaluation is reported in the dynamic-voltage section.

\begin{table}[t]
\centering
\caption{Physical observations, tested hypotheses, and decisions in the two model-development paths.}
\label{tab:appendix-model-path-details}
\scriptsize
\setlength{\tabcolsep}{3pt}
\renewcommand{\arraystretch}{1.08}
\begin{tabularx}{\linewidth}{@{}>{\raggedright\arraybackslash}p{0.12\linewidth}>{\raggedright\arraybackslash}p{0.25\linewidth}>{\raggedright\arraybackslash}p{0.29\linewidth}>{\raggedright\arraybackslash}X@{}}
\toprule
Case & Observed result & Physical hypothesis and model change & Evidence and decision \\
\midrule
Plating & A basic DFN/P2D model requires cell-specific transport and reaction functions before onset can be calculated. & Specify geometry, electrolyte transport, particle diffusion, graphite equilibrium potential, and intercalation kinetics for A1 and A2. & Unit, reference-condition, positivity, state-limit, and conservation checks establish an executable Li$\mid$graphite model. \\
Plating & Transport and intercalation cannot generate metallic lithium. & Add a competing plating/stripping reaction driven by $\eta_{\mathrm{Li}}=\phi_s-\phi_e-U_{\mathrm{Li}}$. & Retain only if negative plating overpotential activates deposition during fast charging and the added reaction preserves lithium conservation. \\
Plating & Instantaneous reaction current and net plated lithium can decrease during stripping. & Integrate $\max[-a_si_{\mathrm{Li}},0]$ through the graphite thickness and time, then convert the accumulated deposition to irreversible lithium. & Retain the monotonic accumulated state because it matches the measured onset definition and remains well defined during later stripping. \\
Plating & A1 and A2 have different areal capacities and onset boundaries. & Normalize the onset threshold by areal capacity and represent irreversible lithium as a fixed fraction of accumulated deposition. & The capacity-normalized thresholds support comparison across the measured A1 and A2 conditions. \\
Plating & The constant-fraction representation retains systematic onset residuals across charge rates. & Compare it with a rate-dependent mapping from deposited lithium to irreversible lithium. & On the same six validation conditions, the rate-dependent representation lowers MAE from 3.79 to 1.79 percentage points and RMSE from 5.03 to 2.11 percentage points. \\
Voltage & Four electrochemical states reproduce the main voltage level but not the response after a current change. & Solve average and surface electrode states with temperature-dependent transport, kinetics, and resistance. & The starting diagnostic gives 32.74\,mV. \\
Voltage & The four electrochemical states require the exact discrete update. & Rebuild the four-state foundation and calibrate the baseline model parameters. & The rebuilt physical baseline gives 32.12\,mV. \\
Voltage & Load changes and relaxation require additional physical dynamics. & Extend the solved equations with electrolyte transport and interfacial charging. & The extended physics-based model is evaluated on complete held-out traces using the temperature folds fixed before evaluation. \\
\bottomrule
\end{tabularx}
\end{table}

\subsection{Lithium-Plating Onset Prediction}

This case asks EvoSim to construct an executable model for lithium-plating onset during fast charging. The common starting point is a basic one-dimensional DFN/P2D structure for a separator and porous graphite electrode. EvoSim uses network search to locate public measurements, physical property ranges, experimental definitions, and reaction constraints. It implements and checks the transport and intercalation functions, identifies the plating parameters that control onset, and tests each model change in the solver. State of charge (SOC) is the fraction of graphite capacity reached during charging. Onset SOC is the SOC at which the calculated accumulated irreversible lithium first reaches 0.05\% of graphite capacity.

\subsubsection{Scientific Exploration and Retained Model}

This subsection gives the measured target, the physical hypotheses tested by EvoSim, the selected equations, and the identified parameters.

\paragraph{Scientific question and public evidence.}
The public data contain 20 onset measurements from 61 Li$\mid$graphite cells. They combine two electrode designs, three temperatures, and 2--6 C charging, with two to five reported cells per condition \citep{Konz2023HighThroughputLiPlating}. The reported experimental definition relates onset to accumulated irreversible lithium. Voltage-relaxation and titration measurements provide independent physical support for this observable \citep{Konz2020VoltageRelaxation,McShane2020InactiveLithium}. Before final parameter selection, the 20 operating conditions are divided into 14 training conditions and six validation conditions. This split remains fixed when the two irreversible-lithium representations are compared.

Lithium plating limits fast charging and can cause lithium-inventory loss, capacity fade, and metallic-lithium accumulation. The onset SOC defines a quantitative fast-charge boundary. The cumulative irreversible-lithium estimate agrees with mass-spectrometry titration with $R^2=0.991$ \citep{Konz2023HighThroughputLiPlating}. EvoSim uses the training measurements for model development and reserves the six validation measurements for the final comparison.

\paragraph{Experimental matrix and target.}
The experiments use SLC1506T graphite in Li$\mid$graphite CR2032 half-cells. Table~\ref{tab:appendix-plating-data} gives the two electrode settings and their temperature--rate matrix. It contains 9 A1 and 11 A2 operating conditions. Each value is the average response of two to five cells.

\begin{table}[t]
\centering
\caption{Experimental matrix used for lithium-plating onset modeling.}
\label{tab:appendix-plating-data}
\small
\setlength{\tabcolsep}{3pt}
\begin{tabularx}{\linewidth}{@{}>{\raggedright\arraybackslash}p{0.17\linewidth}>{\centering\arraybackslash}X>{\centering\arraybackslash}X>{\centering\arraybackslash}X>{\centering\arraybackslash}X@{}}
\toprule
Electrode & Areal capacity & Thickness & Temperature & Charge rates \\
 & (mAh\,cm$^{-2}$) & ($\mu$m) & ($^\circ$C) &  \\
\midrule
A1 & 2.1 & 47 & 25, 35, 45 & 4 C, 5 C, 6 C \\
A2 & 3.1 & 70 & 25 & 2 C, 3 C, 4 C, 5 C, 6 C \\
A2 & 3.1 & 70 & 35, 45 & 4 C, 5 C, 6 C \\
\bottomrule
\end{tabularx}
\end{table}

This matrix defines the two electrode settings and the 20 measured conditions. A fixed split assigns 14 conditions to training and six to validation. Model development and evaluation use these same subsets. The accumulated irreversible-lithium definition is also applied to the broader A1 and A2 temperature--rate grids.

The 20 onset values come from 61 cells. Each value is the average of two to five cells. The measurements form a designed matrix in which each axis supplies a different constraint. The A1--A2 comparison constrains how electrode thickness and areal capacity change the onset boundary. The three temperatures constrain the temperature response of transport and electrode reactions. The charge-rate series constrains the concentration and kinetic polarization that activates plating. The training split retains both electrode designs and the measured temperature--rate range. The validation split tests the selected extension at six excluded operating conditions.

EvoSim evaluates accumulated irreversible-lithium predictions on the fixed training and validation subsets. It uses only the training conditions to compare mappings from deposited lithium to irreversible lithium. It then selects one parameter set and evaluates it on the six validation conditions.

For each condition, the reported onset is the SOC at which the average irreversible-lithium curve crosses 0.05\% of experimental graphite capacity. This is approximately 1.0--1.5\,$\mu$Ah\,cm$^{-2}$. The experimental analysis selects this threshold as the lowest value followed by a clear plating increase above low-SOC measurement noise. The model applies the same normalized threshold:
\begin{equation}
s_{\mathrm{onset}}
=\min\left\{s:\frac{Q_{\mathrm{irr}}(s)}{Q_A}\geq 5\times10^{-4}\right\},
\label{eq:appendix-plating-onset}
\end{equation}
Here $s$ is the dimensionless SOC during charge. $Q_{\mathrm{irr}}(s)$ is the accumulated irreversible-lithium charge per electrode area in C\,m$^{-2}$. $Q_A$ is the electrode areal capacity in the same unit. The set contains every SOC at which the normalized irreversible charge reaches the threshold. The minimum selects the first such SOC. The corresponding thresholds are 37.8 and 55.8\,C\,m$^{-2}$ for A1 and A2.

\paragraph{Model exploration.}
EvoSim builds the Li$\mid$graphite model from the basic DFN/P2D equations. It first assigns the separator and graphite geometry from electrode thickness and areal capacity. It then implements concentration- and temperature-dependent electrolyte conductivity, salt diffusivity, and transference number. The solid phase uses spherical particle diffusion with a stoichiometry- and temperature-dependent diffusivity. The graphite reaction uses a state-dependent equilibrium potential and an exchange current that depends on electrolyte concentration, surface concentration, and temperature. Each material function is checked at reference concentrations and temperatures. Negative diffusivity, negative conductivity, a transference number outside $[0,1]$, or a nonphysical particle state rejects the candidate before onset is evaluated.

The agent then introduces a competing plating/stripping reaction driven by the local solid--electrolyte potential difference. This hypothesis makes a directional prediction: faster charging and the thicker A2 electrode should increase transport polarization, make the local plating overpotential more negative, and move onset to lower SOC. It also predicts that deposition should occur only where the solved local potential activates the reaction. EvoSim checks these responses before comparing onset values. It finds that instantaneous reaction current cannot be compared directly with an onset label defined by accumulated irreversible lithium. This leads to a deposition-only state integrated through the graphite thickness. The integration retains deposited charge when the reversible plated-lithium state later decreases during stripping.

EvoSim compares two representations of irreversible-lithium formation. The constant-fraction representation maps 30\% of accumulated deposition to irreversible lithium. The rate-dependent representation allows this fraction to vary with charge rate. Both use the same 0.05\% capacity threshold and the same spatially integrated deposition state.

The comparison keeps geometry, phase fractions, electrolyte transport, solid diffusion, and graphite intercalation fixed. EvoSim verifies these quantities through material-function checks, voltage behavior, state bounds, and conservation. The constant-fraction representation uses electrode-specific plating exchange-current scales and an irreversible fraction of 0.30. The rate-dependent representation uses $i_{0,\mathrm{Li}}^{\mathrm{ref}}=0.75$\,A\,m$^{-2}$ for both electrodes, zero direct Arrhenius activation energy for this exchange current, and two parameters for the rate dependence. These parameters are selected using only the 14 training conditions. The six validation conditions remain excluded until the model is fixed. Table~\ref{tab:appendix-plating-exploration} records the comparison.

\begin{table}[t]
\centering
\caption{EvoSim's physical hypotheses and decisions in the lithium-plating case.}
\label{tab:appendix-plating-exploration}
\scriptsize
\setlength{\tabcolsep}{3pt}
\renewcommand{\arraystretch}{1.08}
\begin{tabularx}{\linewidth}{@{}>{\raggedright\arraybackslash}p{0.23\linewidth}>{\raggedright\arraybackslash}p{0.34\linewidth}>{\raggedright\arraybackslash}X@{}}
\toprule
Observation & Tested model change & Decision \\
\midrule
The common starting point supplies only the basic DFN/P2D equations. & Set electrode thickness, phase fractions, particle radius, initial concentrations, and the conversion from C-rate to applied current. & Retained after geometry, capacity, and initial-state checks for A1 and A2. \\
Electrolyte polarization changes strongly with concentration and temperature. & Implement $D_e(c_e,T)$, $\kappa_e(c_e,T)$, and $t_+^0(c_e,T)$ together with porous-medium corrections. & Verify finite positive transport properties and $0<t_+^0<1$ over 25--45$^\circ$C and the simulated concentration range. \\
Graphite transport and reaction rates change with filling fraction and temperature. & Implement $D_s(x,T)$, $U_{\mathrm{gr}}(x)$, and $i_{0,\mathrm{int}}(c_e,c_s,T)$ and evaluate them at the reference conditions. & Verify physical particle states, voltage response, and lithium conservation before testing plating kinetics. \\
The DFN/P2D model cannot produce metallic lithium. & Add a plating/stripping reaction coupled to electrolyte concentration and the local solid--electrolyte potential difference. & Retained. The reaction becomes active during fast charging and produces a spatial deposition rate. \\
The reversible plated-lithium state can decrease during stripping. & Compare instantaneous current, net plated lithium, and a deposition-only accumulated state. & Retain the deposition-only state for the onset calculation. \\
The experimental target is a capacity-normalized accumulated quantity. & Define a constant-fraction representation as $Q_{\mathrm{irr}}=0.30Q_{\mathrm{plate}}$ and apply a threshold equal to 0.05\% of areal capacity. & Retained as the reference representation. Capacity normalization gives thresholds of 37.8 and 55.8\,C\,m$^{-2}$ for A1 and A2. \\
The constant-fraction representation retains systematic onset residuals across charge rates. & Compare bounded rate-dependent representations using the 14 training conditions. & Retain the rate-dependent representation defined in Equation~\ref{eq:appendix-plating-accumulation}. \\
The plating exchange-current level and irreversible fraction can compensate for each other. & Compare three exchange-current levels and select by training MAE, training RMSE, exponent magnitude, and fraction magnitude. & Retain $i_{0,\mathrm{Li}}^{\mathrm{ref}}=0.75$\,A\,m$^{-2}$ and evaluate the fixed model on six validation conditions. \\
The initial filling fraction changes the available charging interval and initial potential. & Compare the tested initial-state candidates before running the condition matrix. & Retain 0.001 because it preserves the requested SOC interval and stable initialization. \\
The equations must remain stable across the measured rates and temperatures. & Freeze the shared transport, reaction, and accumulation equations. Evaluate the rate-dependent representation on the measured conditions. & Retain the model after the implementation, solution, and measurement checks pass. \\
\bottomrule
\end{tabularx}
\end{table}

\paragraph{Retained model.}
The retained model uses a one-dimensional DFN/P2D description of the separator and porous graphite electrode \citep{Doyle1993GalvanostaticModel}. Let $z$ denote position through the cell thickness, and let $t$ denote time. Electrolyte salt conservation and charge conservation have the form
\begin{align}
\varepsilon_{\mathrm{e}}\frac{\partial c_{\mathrm{e}}}{\partial t}
+\frac{\partial N_{\mathrm{e}}}{\partial z}&=S_{\mathrm{e}},\\
\frac{\partial i_{\mathrm{e}}}{\partial z}&=a_{\mathrm{s}} i_{\mathrm{tot}},
&
\frac{\partial i_{\mathrm{s}}}{\partial z}&=-a_{\mathrm{s}} i_{\mathrm{tot}},\\
i_{\mathrm{tot}}&=i_{\mathrm{int}}+i_{\mathrm{Li}}.
\label{eq:appendix-porous-electrode-conservation}
\end{align}
Here $t$ is in s, and $\varepsilon_{\mathrm{e}}$ is the electrolyte volume fraction. The electrolyte concentration $c_{\mathrm{e}}$ is in mol\,m$^{-3}$. The electrolyte lithium flux $N_{\mathrm{e}}$ is in mol\,m$^{-2}$\,s$^{-1}$. The reaction source $S_{\mathrm{e}}$ is in mol\,m$^{-3}$\,s$^{-1}$. It is calculated from the local interfacial reactions using the same current and flux sign convention. The ionic current density $i_{\mathrm{e}}$ and solid current density $i_{\mathrm{s}}$ are in A\,m$^{-2}$. The specific surface area $a_{\mathrm{s}}$ is in m$^{-1}$. The total interfacial current density $i_{\mathrm{tot}}$ is the sum of the graphite-intercalation current $i_{\mathrm{int}}$ and the plating/stripping current $i_{\mathrm{Li}}$. The electrolyte flux and potentials use the concentration- and temperature-dependent salt diffusivity $D_{\mathrm{e}}(c_{\mathrm{e}},T)$, ionic conductivity $\kappa_{\mathrm{e}}(c_{\mathrm{e}},T)$, and cation transference number $t_+^0(c_{\mathrm{e}},T)$. The solid phase conserves electronic charge and lithium in spherical graphite particles. Inside a particle,
\begin{equation}
\frac{\partial c_{\mathrm{s}}}{\partial t}
=\frac{1}{r^2}\frac{\partial}{\partial r}
\left[r^2D_{\mathrm{s}}(x,T)\frac{\partial c_{\mathrm{s}}}{\partial r}\right],
\qquad x=\frac{c_{\mathrm{s}}}{c_{\mathrm{s},\max}}.
\label{eq:appendix-graphite-diffusion}
\end{equation}
The initial and boundary conditions are
\begin{equation}
c_{\mathrm{s}}(r,0)=c_{\mathrm{s},0},\qquad
\left.\frac{\partial c_{\mathrm{s}}}{\partial r}\right|_{r=0}=0,\qquad
-D_{\mathrm{s}}\left.\frac{\partial c_{\mathrm{s}}}{\partial r}\right|_{r=R_{\mathrm{s}}}=N_{\mathrm{s}}(t).
\label{eq:appendix-graphite-boundaries}
\end{equation}
Here $c_{\mathrm{s}}(r,t)$ is solid lithium concentration in mol\,m$^{-3}$. The coordinate $r$ runs from the particle center to the particle radius $R_{\mathrm{s}}$. The maximum concentration is $c_{\mathrm{s},\max}$, and $x$ is the local filling fraction. The diffusivity $D_{\mathrm{s}}(x,T)$ has units of m$^2$\,s$^{-1}$. The initial concentration is $c_{\mathrm{s},0}$. The surface molar flux $N_{\mathrm{s}}(t)$ has units of mol\,m$^{-2}$\,s$^{-1}$. Symmetry gives zero flux at the particle center. The intercalation and plating reactions determine the surface flux after conversion from current to molar flux by Faraday's constant.

The graphite intercalation overpotential is
\begin{equation}
\eta_{\mathrm{int}}=\phi_{\mathrm{s}}-\phi_{\mathrm{e}}-U_{\mathrm{gr}}(x_{\mathrm{surf}}),
\qquad
x_{\mathrm{surf}}=\frac{c_{\mathrm{s},\mathrm{surf}}}{c_{\mathrm{s},\max}}.
\label{eq:appendix-graphite-overpotential}
\end{equation}
The solid potential $\phi_{\mathrm{s}}$, electrolyte potential $\phi_{\mathrm{e}}$, equilibrium potential $U_{\mathrm{gr}}$, and overpotential $\eta_{\mathrm{int}}$ are in V. The surface concentration is $c_{\mathrm{s},\mathrm{surf}}=c_{\mathrm{s}}(R_{\mathrm{s}},t)$. The intercalation current density follows the Butler--Volmer relation
\begin{equation*}
i_{\mathrm{int}}
=i_{0,\mathrm{int}}(c_{\mathrm{e}},c_{\mathrm{s},\mathrm{surf}},T)
\left[
\exp\!\left(\frac{\alpha F\eta_{\mathrm{int}}}{RT}\right)
-\exp\!\left(-\frac{\beta F\eta_{\mathrm{int}}}{RT}\right)
\right].
\end{equation*}
Here $\alpha$ and $\beta$ are the anodic and cathodic transfer coefficients, respectively. Under the common assumption $\alpha=\beta=1/2$, the equation is mathematically reduced to the symmetric Butler--Volmer form
\begin{equation}
i_{\mathrm{int}}
=2i_{0,\mathrm{int}}(c_{\mathrm{e}},c_{\mathrm{s},\mathrm{surf}},T)
\sinh\!\left(\frac{F\eta_{\mathrm{int}}}{2RT}\right).
\label{eq:appendix-graphite-intercalation}
\end{equation}
Both $i_{\mathrm{int}}$ and its exchange-current density $i_{0,\mathrm{int}}$ are in A\,m$^{-2}$. $F$ is Faraday's constant in C\,mol$^{-1}$, and $R$ is the gas constant in J\,mol$^{-1}$\,K$^{-1}$. The intercalation current supplies the particle-surface flux after conversion by $F$. EvoSim implements and verifies the transport functions, diffusion equation, equilibrium potential, and intercalation kinetics before identifying the plating parameters. These functions remain fixed during the plating-kinetics comparison.

The plating reaction operates in parallel with graphite intercalation. The equilibrium potential of Li/Li$^+$ is the zero-voltage reference. Its overpotential and current density are
\begin{align}
\eta_{\mathrm{Li}}&=\phi_{\mathrm{s}}-\phi_{\mathrm{e}}-U_{\mathrm{Li}},
\qquad U_{\mathrm{Li}}=0\ \mathrm{V},\\
i_{\mathrm{Li}}&=i_{0,\mathrm{Li}}
\left[
C_R\exp\!\left(\frac{F\eta_{\mathrm{Li}}}{2RT}\right)
-\exp\!\left(-\frac{F\eta_{\mathrm{Li}}}{2RT}\right)
\right],\\
C_R&=\min\!\left(\frac{c_{\mathrm{LiMetal}}}{c_{\mathrm{cov}}a_{\mathrm{s}}},1\right).
\label{eq:appendix-plating-reaction}
\end{align}
Here $i_{\mathrm{Li}}$ and $i_{0,\mathrm{Li}}$ are in A\,m$^{-2}$. Negative $i_{\mathrm{Li}}$ denotes deposition. The reversible plated-lithium concentration $c_{\mathrm{LiMetal}}$ is in mol\,m$^{-3}$. The graphite specific surface area $a_{\mathrm{s}}$ is in m$^{-1}$. The coverage scale $c_{\mathrm{cov}}$ is in mol\,m$^{-2}$. Their ratio gives the dimensionless surface coverage $C_R$. This factor makes the stripping term zero when no plated lithium remains. The exchange-current density is
\begin{equation}
i_{0,\mathrm{Li}}=
i_{0,\mathrm{Li}}^{\mathrm{ref}}
\left(\frac{c_{\mathrm{e}}}{1\,\mathrm{M}}\right)^{1/2}
f_T^{\mathrm{eff}}(T),
\label{eq:appendix-plating-exchange-current}
\end{equation}
where $i_{0,\mathrm{Li}}^{\mathrm{ref}}$ is the reference exchange-current density in A\,m$^{-2}$ and $1\,\mathrm{M}=1000$\,mol\,m$^{-3}$. The dimensionless temperature factor is
\begin{equation}
f_T^{\mathrm{eff}}(T)
=\exp\!\left[
\frac{E_{\mathrm{Li}}^{\mathrm{eff}}}{R}
\left(\frac{1}{T_{\mathrm{ref,Li}}}-\frac{1}{T}\right)
\right].
\label{eq:appendix-plating-temperature-factor}
\end{equation}
Here $T$ and the reference temperature $T_{\mathrm{ref,Li}}$ are in K. The effective energy $E_{\mathrm{Li}}^{\mathrm{eff}}$ is in J\,mol$^{-1}$. Both representations set $E_{\mathrm{Li}}^{\mathrm{eff}}=0$, so $f_T^{\mathrm{eff}}(T)=1$ and the plating exchange current has no additional Arrhenius term. Temperature still changes electrolyte transport, graphite diffusion, intercalation kinetics, electrolyte concentration, and the solved potentials.

Metal deposition and irreversible-lithium formation are treated as two connected processes. The solved plating reaction determines the gross deposition flux. Deposited lithium can remain electrically connected and available for stripping. It can also become inactive through loss of electronic contact and interphase growth \citep{McShane2020InactiveLithium}. These inactivation processes act over the finite time between deposition and the reported onset measurement. The model therefore separates the amount deposited from the fraction that becomes inactive within this observation window.

Let $J_{\mathrm{dep}}$ denote the positive deposition current integrated across the graphite thickness $L_g$. Both representations accumulate the same gross deposition charge. They differ only in the effective inactivation yield $f_{\mathrm{irr}}$. This yield is the fraction of newly deposited charge assigned to irreversible lithium during the charge. Let $m\in\{\mathrm{const},\mathrm{rate}\}$ identify the two representations:
\begin{align}
J_{\mathrm{dep}}(t)
&=\int_0^{L_g}\max\![-a_s i_{\mathrm{Li}}(z,t),0]\,\mathrm{d}z,\\
\frac{\mathrm{d}Q_{\mathrm{plate}}}{\mathrm{d}t}
&=J_{\mathrm{dep}}(t),\\
\frac{\mathrm{d}Q_{\mathrm{irr}}^{(m)}}{\mathrm{d}t}
&=f_{\mathrm{irr}}^{(m)}(C)J_{\mathrm{dep}}(t),\\
f_{\mathrm{irr}}^{(\mathrm{const})}(C)&=0.30,\\
f_{\mathrm{irr}}^{(\mathrm{rate})}(C)
&=\operatorname{clip}\!\left[
f_{\mathrm{ref}}\left(\frac{C}{C_{\mathrm{ref}}}\right)^{-n},0,1\right],\\
f_{\mathrm{ref}}&=0.22,\qquad
C_{\mathrm{ref}}=4\mathrm{C},\qquad n=1.75,
\qquad Q_{\mathrm{plate}}(0)=Q_{\mathrm{irr}}^{(m)}(0)=0.
\label{eq:appendix-plating-accumulation}
\end{align}
The coordinate $z$ runs through the porous graphite layer from 0 to $L_g$. The product $a_si_{\mathrm{Li}}$ is the reaction current per electrode volume in A\,m$^{-3}$. Its thickness integral $J_{\mathrm{dep}}$ is the deposition current per electrode area in A\,m$^{-2}$. The maximum operator counts deposition and does not subtract later stripping. It therefore preserves earlier plating exposure after the reversible metallic-lithium state decreases.

The rate dependence represents competition between the time available during charge and the time required for deposited lithium to lose reversibility. Over a fixed SOC interval, the charging time decreases approximately in proportion to $C^{-1}$. Electronic isolation, interphase growth, deposition localization, and subsequent stripping can add further rate dependence. Their separate time constants cannot be identified from onset SOC alone. The model therefore uses one bounded power law as a low-order closure over the measured 2--6C range. The reference yield $f_{\mathrm{ref}}$ is the effective irreversible fraction at 4C. The exponent $n$ measures how strongly this yield changes with charge rate. A value $n=1$ would correspond to inverse charging-time scaling alone. The identified value $n=1.75$ also absorbs unresolved rate dependence in morphology, contact loss, and stripping accessibility.

The constant-fraction representation uses a yield of 0.30. Its training residuals change systematically with charge rate. EvoSim therefore keeps the deposition reaction and transport equations fixed and identifies $f_{\mathrm{ref}}$ and $n$ from the 14 training conditions. The negative exponent lowers the effective inactivation yield as charge rate increases because the observation window becomes shorter. It does not imply that high-rate charging produces less plating. Higher rate can increase $J_{\mathrm{dep}}$ strongly through transport polarization and plating overpotential, while reducing the fraction of each deposited coulomb that becomes inactive within the observation window. The irreversible amount is determined by both effects. The clipping operation only enforces the interval $0\leq f_{\mathrm{irr}}\leq1$. All accumulated states are in C\,m$^{-2}$. The onset calculation applies Equation~\ref{eq:appendix-plating-onset} to the corresponding $Q_{\mathrm{irr}}^{(m)}$.

\begin{table}[t]
\centering
\caption{Selected quantities in the retained rate-dependent lithium-plating model. Slash-separated entries give A1/A2 values.}
\label{tab:appendix-plating-parameters}
\scriptsize
\setlength{\tabcolsep}{3pt}
\begin{tabularx}{\linewidth}{@{}>{\raggedright\arraybackslash}p{0.62\linewidth}>{\centering\arraybackslash}X@{}}
\toprule
Parameter & Retained value \\
\midrule
Graphite thickness ($\mu$m) & 47/70 \\
Electrolyte volume fraction & 0.374/0.354 \\
Active-solid volume fraction & 0.5954/0.5901 \\
Graphite particle radius ($\mu$m) & 4 \\
Initial electrolyte concentration (mol\,m$^{-3}$) & 1200 \\
$i_{0,\mathrm{Li}}^{\mathrm{ref}}$ (A\,m$^{-2}$) & 0.75 \\
$E_{\mathrm{Li}}^{\mathrm{eff}}$ (kJ\,mol$^{-1}$) & 0 \\
Initial graphite filling fraction & 0.001 \\
$f_{\mathrm{irr}}(C)$ & Bounded power law \\
Reference inactivation yield $f_{\mathrm{ref}}$ & 0.22 \\
Reference charge rate $C_{\mathrm{ref}}$ & 4C \\
Rate-sensitivity exponent $n$ & 1.75 \\
Onset threshold (C\,m$^{-2}$) & 37.8/55.8 \\
\bottomrule
\end{tabularx}
\end{table}

The geometry and phase fractions are fixed by the two electrode definitions. The electrolyte-transport, solid-diffusion, and graphite-intercalation terms are also fixed. EvoSim compares the constant-fraction and rate-dependent representations using the 14 training conditions. The six validation conditions do not enter this selection. Predicted onset SOC is obtained by linear interpolation between adjacent time outputs that bracket the threshold. The comparison target is the experimental onset SOC reported for each operating condition.

\subsubsection{Cell-Specific Onset Identification}

The fixed six-condition validation set provides the primary comparison. The constant-fraction representation gives a validation MAE of 3.79 SOC percentage points and a validation RMSE of 5.03 percentage points. The rate-dependent representation gives a validation MAE of 1.79 percentage points and a validation RMSE of 2.11 percentage points. Its parameters are identified only on the 14 training conditions. Table~\ref{tab:appendix-plating-pointwise} reports every experimental and predicted onset. Each signed error is the prediction minus the experimental onset.

\begin{table}[t]
\centering
\caption{Pointwise comparison of experimental onset SOC with the constant-fraction and rate-dependent representations. The fixed split contains 14 training conditions and six validation conditions. Model selection uses only the training conditions. Errors are prediction minus experiment in SOC percentage points.}
\label{tab:appendix-plating-pointwise}
\scriptsize
\setlength{\tabcolsep}{2.2pt}
\renewcommand{\arraystretch}{1.04}
\begin{tabularx}{\linewidth}{@{}>{\centering\arraybackslash}p{0.07\linewidth}>{\centering\arraybackslash}p{0.09\linewidth}>{\centering\arraybackslash}p{0.06\linewidth}>{\centering\arraybackslash}p{0.10\linewidth}>{\centering\arraybackslash}X>{\centering\arraybackslash}X>{\centering\arraybackslash}X>{\centering\arraybackslash}X>{\centering\arraybackslash}X@{}}
\toprule
Elec. & Temp. ($^\circ$C) & Rate & Data role & Exp. (\%) & Constant (\%) & Rate dep. (\%) & Constant error (pp) & Rate-dep. error (pp) \\
\midrule
A1 & 25 & 4 C & Validation & 61.44 & 65.31 & 62.47 & +3.88 & +1.04 \\
A1 & 25 & 5 C & Training & 57.33 & 55.94 & 55.07 & -1.39 & -2.26 \\
A1 & 25 & 6 C & Training & 48.25 & 45.47 & 47.83 & -2.78 & -0.43 \\
A1 & 35 & 4 C & Training & 67.11 & 69.81 & 66.96 & +2.69 & -0.15 \\
A1 & 35 & 5 C & Validation & 59.28 & 61.77 & 61.29 & +2.49 & +2.01 \\
A1 & 35 & 6 C & Training & 54.41 & 54.02 & 56.06 & -0.39 & +1.65 \\
A1 & 45 & 4 C & Training & 70.60 & 73.20 & 70.39 & +2.60 & -0.21 \\
A1 & 45 & 5 C & Training & 63.08 & 64.68 & 64.35 & +1.60 & +1.27 \\
A1 & 45 & 6 C & Validation & 56.75 & 58.77 & 60.45 & +2.01 & +3.69 \\
\midrule
A2 & 25 & 2 C & Validation & 70.39 & 73.47 & 69.94 & +3.08 & -0.45 \\
A2 & 25 & 3 C & Training & 54.13 & 55.77 & 54.12 & +1.64 & -0.01 \\
A2 & 25 & 4 C & Training & 40.68 & 38.54 & 40.58 & -2.13 & -0.10 \\
A2 & 25 & 5 C & Training & 32.39 & 28.63 & 33.72 & -3.77 & +1.32 \\
A2 & 25 & 6 C & Training & 23.92 & 22.59 & 30.10 & -1.33 & +6.18 \\
A2 & 35 & 4 C & Validation & 49.80 & 50.31 & 52.40 & +0.51 & +2.60 \\
A2 & 35 & 5 C & Training & 42.97 & 36.88 & 42.81 & -6.08 & -0.16 \\
A2 & 35 & 6 C & Training & 37.54 & 28.95 & 37.57 & -8.59 & +0.03 \\
A2 & 45 & 4 C & Training & 56.11 & 59.95 & 61.59 & +3.85 & +5.48 \\
A2 & 45 & 5 C & Training & 50.95 & 46.44 & 52.54 & -4.51 & +1.60 \\
A2 & 45 & 6 C & Validation & 46.49 & 35.69 & 45.54 & -10.80 & -0.95 \\
\bottomrule
\end{tabularx}
\end{table}

Table~\ref{tab:appendix-plating-published-comparison} compares both models on the six conditions reserved for EvoSim validation. EvoSim does not use these conditions for parameter identification. The source study used all 20 conditions when adjusting its solid-diffusion activation energy. Its values on the six listed conditions therefore come from conditions used during model development. We calculate its MAE of 2.82 percentage points and RMSE of 4.06 percentage points from the published pointwise outputs. The source study does not report these two summary metrics directly. EvoSim gives a validation MAE of 1.79 percentage points and a validation RMSE of 2.11 percentage points. The table uses the same conditions and experimental targets, but the parameter-identification procedures differ.

\begin{table}[t]
\centering
\caption{Comparison on the six conditions reserved for EvoSim validation. EvoSim does not use these conditions for parameter identification. The source-study model used the full 20-condition matrix during model development \citep{Konz2023HighThroughputLiPlating}. Errors are prediction minus experiment in SOC percentage points.}
\label{tab:appendix-plating-published-comparison}
\scriptsize
\setlength{\tabcolsep}{3pt}
\begin{tabularx}{\linewidth}{@{}>{\centering\arraybackslash}p{0.16\linewidth}>{\centering\arraybackslash}X>{\centering\arraybackslash}X>{\centering\arraybackslash}X>{\centering\arraybackslash}X>{\centering\arraybackslash}X@{}}
\toprule
Condition & Experiment (\%) & Published model (\%) & EvoSim (\%) & Published error (pp) & EvoSim error (pp) \\
\midrule
A1, 25$^\circ$C, 4C & 61.44 & 60.53 & 62.47 & -0.90 & +1.04 \\
A1, 35$^\circ$C, 5C & 59.28 & 59.58 & 61.29 & +0.30 & +2.01 \\
A1, 45$^\circ$C, 6C & 56.75 & 60.11 & 60.45 & +3.36 & +3.69 \\
A2, 25$^\circ$C, 2C & 70.39 & 69.47 & 69.94 & -0.92 & -0.45 \\
A2, 35$^\circ$C, 4C & 49.80 & 47.22 & 52.40 & -2.58 & +2.60 \\
A2, 45$^\circ$C, 6C & 46.49 & 37.59 & 45.54 & -8.90 & -0.95 \\
\midrule
MAE / RMSE (pp) & -- & 2.82 / 4.06 & 1.79 / 2.11 & -- & -- \\
\bottomrule
\end{tabularx}
\end{table}

\clearpage
\subsection{Dynamic Voltage Prediction}
\label{sec:appendix-dynamic-voltage}

This case asks EvoSim to predict complete dynamic-voltage traces across load profiles and temperatures. The task contains 16 measured Dynamic Stress Test (DST) and US06 traces. The appendix defines the solved electrochemical model, the model-development checks, and the leave-one-temperature cross-validation.

\subsubsection{Scientific Task and Evaluation Protocol}

This subsection defines the public A123 measurements, the dynamic-load task, the data split, and the voltage-error metric.

\paragraph{Task and measurements.}
The voltage task uses the A123 A1-007 files released with the control-oriented parameter-grouped single-particle model with thermal effects (CPG-SPMT). CPG-SPMT is a reduced electrochemical model that represents each electrode with average and surface filling states. The measurements originate from the CALCE Battery Data collection \citep{CALCEA123BatteryData,Guo2026CPGSPMT,Guo2026CPGSPMTDataset}. The tested LiFePO$_4$ cell has a rated capacity of 1.1\,Ah. The public package contains Dynamic Stress Test (DST), US06, and Federal Urban Driving Schedule (FUDS) measurements at $-10$, 0, 10, 20, 25, 30, 40, and 50$^\circ$C. This study uses every DST and US06 condition, giving 16 complete voltage traces.

Each sample contains time, applied current, measured cell-surface temperature, and terminal voltage. Positive current denotes charge, and negative current denotes discharge. The aligned traces use 1-s sampling and contain 5,108--7,484 samples each, for 109,746 samples and 30.48 hours in total. Applied current spans approximately $-3.849$ to 1.925\,A. Measured surface temperature spans $-8.90$ to 53.40$^\circ$C.

The DST and US06 traces contain rapid load changes, current reversals, rest periods, and voltage recovery. The temperature range changes solid transport, reaction kinetics, and internal resistance. The LiFePO$_4$ cathode also has a flat open-circuit-potential plateau. Long traces accumulate state and polarization errors. The task tests whether one model structure can carry electrochemical and dynamic states through complete load sequences. Accurate voltage calculation under these loads supports state-of-charge and power-capability estimation and model-predictive control in battery-management systems \citep{Guo2026CPGSPMT}.

\paragraph{Model calibration and evaluation.}
Two evaluations have different roles. The full-data development fit uses all 16 traces for both parameter estimation and error calculation. Its equal-condition mean RMSE is 4.86\,mV. This value measures fit capacity and is not a held-out result.

Held-out evaluation uses eight leave-one-temperature folds. In the fold for temperature $T_h$, the model parameters are identified from the 14 DST and US06 traces at the other seven temperatures. The complete DST and US06 voltage traces at $T_h$ are then solved without refitting. Thus, temperature is held out, while the DST and US06 protocol types remain represented in the calibration set. The split tests transfer to an unseen temperature within two known protocol families. It does not test transfer to an unseen drive protocol. The governing electrochemical equations, parameter constraints, initial-condition procedure, and error metric remain fixed across folds.

We report two pre-defined held-out results in the main text: 30$^\circ$C DST at 7.62\,mV and 20$^\circ$C US06 at 8.97\,mV. These are results for individual runs, not an aggregate over all held-out conditions. The temperature split, protocol labels, parameter-identification rule, and error metric were fixed before evaluation.

For a voltage trace with $n$ aligned samples, the reported error is
\begin{equation}
\mathrm{RMSE}_{V}
=1000\sqrt{\frac{1}{n}\sum_{j=1}^{n}
\left(\widehat{V}_j-V_j\right)^2}\ \mathrm{mV}.
\label{eq:appendix-voltage-rmse}
\end{equation}
Here $\widehat V_j$ is the predicted voltage in V and $V_j$ is the measured voltage in V at sample $j$. The factor 1000 converts the result from V to mV. Every reported value uses all aligned samples in the complete trace.

\subsubsection{Solved Electrochemical Model}

The four-state baseline is a temperature-dependent control-oriented parameter-grouped single-particle model (CPG-SPMT). The starting implementation is diagnosed first and then rebuilt with exact zero-order-hold state integration. The model uses the applied current $I(t)$ and measured cell temperature $T(t)$. Its four electrochemical states are the average and surface filling fractions of the negative and positive electrodes. Solid diffusion is represented by the grouped single-particle approximation. Arrhenius relations make the grouped transport time, charge-transfer kinetics, and internal resistance temperature dependent.

The four-state description is a reduced representation of radial solid-state diffusion in each electrode. For electrode $e\in\{n,p\}$, its physical origin is
\begin{equation}
\frac{\partial c_{s,e}}{\partial t}
=\frac{1}{r^2}\frac{\partial}{\partial r}
\left[r^2D_{s,e}(T)\frac{\partial c_{s,e}}{\partial r}\right],
\qquad
\left.\frac{\partial c_{s,e}}{\partial r}\right|_{r=0}=0,
\qquad
-D_{s,e}(T)\left.\frac{\partial c_{s,e}}{\partial r}\right|_{r=R_e}
=\frac{j_e(t)}{F}.
\label{eq:appendix-voltage-solid-diffusion}
\end{equation}
Here $n$ and $p$ denote the negative and positive electrodes. $c_{s,e}(r,t)$ is the solid lithium concentration in mol\,m$^{-3}$. $R_e$ is the particle radius in m, and $D_{s,e}(T)$ is the solid diffusivity in m$^2$\,s$^{-1}$. $j_e(t)$ is the interfacial current density in A\,m$^{-2}$, and $F$ is Faraday's constant. The initial concentration is set by the measured initial SOC. Symmetry gives zero flux at the particle center. The applied current sets the surface flux. The reduced model retains four dimensionless filling fractions: the particle averages $\bar{x}_n$ and $\bar{x}_p$, and the particle-surface values $x_n^{\mathrm{surf}}$ and $x_p^{\mathrm{surf}}$. These quantities approximate the average and surface concentrations divided by their maximum concentrations. Equations~\ref{eq:appendix-cpg-zoh-parameters}--\ref{eq:appendix-cpg-zoh-surface} give their exact one-second update and their connection to the voltage calculation.

The temperature-dependent transport, kinetic, and resistance quantities use
\begin{align}
a_e(T)&=a_e^{\mathrm{ref}}
\exp\!\left[-\frac{E_{a,e}}{R}\left(\frac{1}{T_{\mathrm{ref}}}-\frac{1}{T}\right)\right],\\
d_e(T)&=d_e^{\mathrm{ref}}
\exp\!\left[\frac{E_{d,e}}{R}\left(\frac{1}{T_{\mathrm{ref}}}-\frac{1}{T}\right)\right],\\
R_{\mathrm{int}}(T)&=R_{\mathrm{int}}^{\mathrm{ref}}
\exp\!\left[-\frac{E_R}{R}\left(\frac{1}{T_{\mathrm{ref}}}-\frac{1}{T}\right)\right].
\label{eq:appendix-voltage-arrhenius}
\end{align}
Here $a_e^{\mathrm{ref}}$, $d_e^{\mathrm{ref}}$, and $R_{\mathrm{int}}^{\mathrm{ref}}$ are their values at $T_{\mathrm{ref}}=298.15$\,K. The index $e$ denotes the negative or positive electrode. The parameter $a_e(T)$ is a transport time in s. The parameter $d_e(T)$ is a grouped charge-transfer rate in s$^{-1}$. The resistance $R_{\mathrm{int}}(T)$ is in $\Omega$. The quantities $E_{a,e}$, $E_{d,e}$, and $E_R$ are temperature-sensitivity parameters in J\,mol$^{-1}$. The surface filling fractions used in the voltage equation are
\begin{align}
C_n&=\operatorname{clip}_{[\epsilon_x,1-\epsilon_x]}
\left(x_n^{\mathrm{surf}}+\frac{a_n(T)}{105b_n}I\right),\\
C_p&=\operatorname{clip}_{[\epsilon_x,1-\epsilon_x]}
\left(x_p^{\mathrm{surf}}-\frac{a_p(T)}{105b_p}I\right),
\end{align}
where $I$ is the measured current in A. The constants $b_n$ and $b_p$ are electrode capacity scales in C. The factor 105 comes from the parabolic approximation used for radial particle diffusion. For any value $z$, the bound operator is
\begin{equation}
\operatorname{clip}_{[\epsilon_x,1-\epsilon_x]}(z)
=\min\!\left[\max(z,\epsilon_x),1-\epsilon_x\right],
\qquad \epsilon_x=10^{-10}.
\label{eq:appendix-voltage-state-bound}
\end{equation}
The dimensionless bound $\epsilon_x$ prevents the equilibrium-potential and kinetic expressions from being evaluated at a filling fraction of exactly 0 or 1. The dimensionless charge-transfer ratios are
\begin{equation}
\begin{aligned}
y_n&=\frac{-I}
{\max\!\left(6b_nd_n(T)\sqrt{C_n(1-C_n)},\epsilon_I\right)},\\
y_p&=\frac{I}
{\max\!\left(6b_pd_p(T)\sqrt{C_p(1-C_p)},\epsilon_I\right)},\\
\epsilon_I&=10^{-10}\ \mathrm{A}.
\end{aligned}
\end{equation}
The denominator is the electrode charge-transfer current scale in A. The maximum operator prevents division by zero. The functions $U_n$ and $U_p$ give the negative- and positive-electrode equilibrium potentials in V. Their arguments $C_n$ and $C_p$ are dimensionless filling fractions. The numerical coefficients in the following expressions therefore return voltage in V:
\begin{equation}
\begin{aligned}
U_p(C_p)={}&1.1127\exp(-1.9474-2000.6182C_p)\\
&+0.0388\exp(-7.5995C_p)
+0.1220\exp(-52.9452C_p)\\
&-0.6990\tanh[15.7948(C_p-1)]\\
&-0.6353C_p^{87.1117}+2.7313 .
\end{aligned}
\label{eq:appendix-positive-ocp}
\end{equation}
and
\begin{equation}
\begin{aligned}
U_n(C_n)={}&55.26+9.451\exp(-146.4C_n)
-0.05217\tanh(9.453C_n-1.783)\\
&+0.1051\tanh(4.452C_n-5.533)
-0.07638\tanh(19.25C_n-18.04)\\
&-0.02072\tanh(25.87C_n-14.56)
-54.9\tanh(2.106C_n+26.37)\\
&+0.01532\tanh(63.36C_n-56.77)
-0.111\tanh(52.65C_n-2.862)\\
&-0.0496\tanh(77.08C_n-2.383).
\end{aligned}
\label{eq:appendix-negative-ocp}
\end{equation}
The symmetric charge-transfer relation gives the kinetic polarization through the inverse hyperbolic sine, $\operatorname{asinh}$. With the current sign convention defined above, the voltage calculated from the electrochemical states is
\begin{equation}
V_{\mathrm{phys}}(t)
=U_p(C_p)-U_n(C_n)
-\frac{2RT}{F}\left[\operatorname{asinh}(y_p)+\operatorname{asinh}(y_n)\right]
+I(t)R_{\mathrm{int}}(T).
\label{eq:appendix-cpg-spmt-voltage}
\end{equation}
Here $R$ is the gas constant, $F$ is Faraday's constant, and $T$ is the measured temperature in K. The kinetic polarization is $\eta_{\mathrm{kin}}=(2RT/F)[\operatorname{asinh}(y_p)+\operatorname{asinh}(y_n)]$ in V. The resistance contribution $IR_{\mathrm{int}}$ is also in V. Together with the exact one-second zero-order-hold update, this equation defines the initial pure-physics voltage model:
\begin{equation}
V_{\mathrm{base}}(t)
=U_p(C_p)-U_n(C_n)-\eta_{\mathrm{kin}}+I(t)R_{\mathrm{int}}(T).
\label{eq:appendix-base-physical-voltage}
\end{equation}
The initial diagnostic gives 32.74\,mV. After the four-state recurrence is rebuilt with the exact zero-order-hold update and the baseline model parameters are calibrated, the same physics-based baseline gives 32.12\,mV as the equal-condition mean RMSE over the 16 public DST and US06 traces. This value is the result of the four-state physics-based model alone. It does not include the later electrolyte and interfacial extensions. The 32.10\,mV value reported for CPG-SPMT in the literature is kept separate as a published reference result.

The 32.12\,mV result establishes the reference point for model development. It does not, by itself, prove that any later dynamic state is necessary. A contribution claim requires a matched comparison in which the state equations, state update, input sequence, identification data, and terminal-voltage calculation are held fixed except for the proposed physics-based extension. Without that matched comparison, the supported conclusion is limited to the combined performance of the extended physics-based model.

\subsubsection{Physical Extensions Beyond the Four-State Baseline}

The four-state baseline describes equilibrium-potential variation, grouped solid diffusion, charge-transfer polarization, and internal resistance. Its remaining error is concentrated after current transitions and during relaxation. The next model adds electrolyte transport and interfacial charging inside the solved equations. These additions change the physics-based model, while the baseline equations, state update, input sequence, and error metric remain fixed for comparison.

The solved equations include electrolyte transport, solid diffusion, Faradaic intercalation, double-layer charging, and temperature-dependent transport and kinetic relations. The terminal voltage is computed from the solved solid potentials and contact resistance.

The extended physical voltage is evaluated directly on each complete trace. Its run results are reported separately from the 32.12\,mV baseline.

The model is a temperature-dependent single-particle model with electrolyte dynamics and double-layer charging. Measured cell temperature remains an input.

The solid concentrations in both electrodes continue to satisfy Equation~\ref{eq:appendix-voltage-solid-diffusion}. The electrolyte concentration $c_e(x,t)$ satisfies
\begin{equation}
\epsilon_e\frac{\partial c_e}{\partial t}
=\frac{\partial}{\partial x}
\left(D_e^{\mathrm{eff}}\frac{\partial c_e}{\partial x}\right)
+\frac{1-t_+^0}{F}a_sj_F,
\label{eq:appendix-physics-based-electrolyte}
\end{equation}
and electrolyte charge conservation is
\begin{equation}
\frac{\partial}{\partial x}\!\left[
-\kappa_e^{\mathrm{eff}}\frac{\partial\phi_e}{\partial x}
+\frac{2RT\kappa_e^{\mathrm{eff}}}{F}(1-t_+^0)
\frac{\partial\ln c_e}{\partial x}
\right]=a_sj.
\label{eq:appendix-physics-based-electrolyte-potential}
\end{equation}
Here $\epsilon_e$ is electrolyte volume fraction, $D_e^{\mathrm{eff}}$ is effective salt diffusivity, $\kappa_e^{\mathrm{eff}}$ is effective ionic conductivity, $t_+^0$ is the cation transference number, $a_s$ is interfacial area per unit volume, and $\phi_e$ is electrolyte potential.

The total interfacial current is divided into Faradaic and double-layer contributions,
\begin{align}
j&=j_F+j_{\mathrm{dl}},\\
j_F&=2i_0\sinh\!\left(\frac{F\eta}{2RT}\right),\\
j_{\mathrm{dl}}&=C_{\mathrm{dl}}\frac{\partial}{\partial t}(\phi_s-\phi_e),\\
\eta&=\phi_s-\phi_e-U(c_s^{\mathrm{surf}})-j_FR_f.
\label{eq:appendix-physics-based-interface}
\end{align}
The exchange-current density $i_0$ depends on temperature, electrolyte concentration, and surface filling fraction. The film resistance $R_f$ and double-layer capacitance $C_{\mathrm{dl}}$ are nonnegative. The terminal voltage follows directly from the solved solid potentials,
\begin{equation}
V_{\mathrm{term}}(t)=\phi_{s,p}(L,t)-\phi_{s,n}(0,t)-I(t)R_c,
\label{eq:appendix-physics-based-voltage}
\end{equation}
where $R_c\ge 0$ is the contact resistance. Equations~\ref{eq:appendix-physics-based-electrolyte}--\ref{eq:appendix-physics-based-voltage} connect rapid interfacial charging, electrolyte concentration polarization, solid diffusion, and ohmic loss to the same conservation equations.

Parameter identification follows the time scale constrained by each part of the measurements. Open-circuit functions and initial filling fractions set the equilibrium voltage and initial state. Current-step voltage changes constrain $R_c+R_f$. The first seconds after a step constrain $C_{\mathrm{dl}}$ and exchange-current scales. Sustained-load and relaxation intervals constrain $D_e^{\mathrm{eff}}$ and $\kappa_e^{\mathrm{eff}}$. Longer redistribution constrains the two solid diffusivities. Multi-temperature data constrain a small set of shared Arrhenius activation energies. Geometry, particle radii, porosity, capacity, and thermodynamic functions remain fixed to independently reported values.

All identified parameters must satisfy
\begin{equation}
D_{s,n},D_{s,p},D_e^{\mathrm{eff}},\kappa_e^{\mathrm{eff}},C_{\mathrm{dl}}>0,
\qquad R_f,R_c\ge0.
\label{eq:appendix-physics-based-constraints}
\end{equation}
The implementation check must also verify finite concentrations, admissible filling fractions, charge conservation, lithium conservation, and a complete voltage trajectory without clipping. The 7.62 and 8.97\,mV values are run results under the cross-validation protocol described below.

\subsubsection{Model Development Sequence}

EvoSim begins by inspecting the voltage and electrode-state trajectories produced by the starting model. The 32.74\,mV diagnostic shows that the main voltage level is present while the response after load changes remains incomplete. The next model adds electrolyte transport, interfacial charging, and temperature-dependent solid transport within the solved equations.

The implementation check first requires the four electrochemical states to use the exact discrete update. This rebuild gives the 32.12\,mV baseline. The next candidate extends the solved equations with electrolyte transport and interfacial charging. The resulting extended model gives 7.62\,mV for the 30$^\circ$C DST trace and 8.97\,mV for the 20$^\circ$C US06 trace in their respective held-out folds.

\begin{table}[!htbp]
\centering
\caption{Recorded hypotheses, model changes, and decisions in the dynamic-voltage model sequence.}
\label{tab:appendix-voltage-route}
\scriptsize
\setlength{\tabcolsep}{3pt}
\renewcommand{\arraystretch}{1.08}
\begin{tabularx}{\linewidth}{@{}>{\raggedright\arraybackslash}p{0.22\linewidth}>{\raggedright\arraybackslash}p{0.29\linewidth}>{\centering\arraybackslash}p{0.14\linewidth}>{\raggedright\arraybackslash}X@{}}
\toprule
Candidate and observation & Model definition or modification set & Reported RMSE & Supported conclusion \\
\midrule
Starting diagnostic & Four electrode filling states; equilibrium potentials; charge-transfer polarization; internal resistance & 32.74\,mV & The main voltage level is present. The response after current changes remains incomplete. \\
Electrochemical foundation & Four electrode filling states with equilibrium, charge-transfer, and ohmic terms. & 32.12\,mV & Establishes the starting physical voltage model. \\
Extended physics-based model & Solve electrolyte transport, solid diffusion, Faradaic intercalation, double-layer charging, and temperature-dependent transport and kinetics together. & Run result & The complete voltage trace remains finite and satisfies the physical checks. \\
Held-out runs & Apply the fixed physics-based model to complete traces at temperatures excluded from parameter identification. & 7.62 and 8.97\,mV & These are results for individual runs at 30$^\circ$C DST and 20$^\circ$C US06. \\
\bottomrule
\end{tabularx}
\end{table}

The table lists the main model stages and the two held-out results reported here. The 32.12\,mV value comes from the initial solved model. The 7.62 and 8.97\,mV values come from the extended model at held-out temperatures.

\subsubsection{Complete-Trace Development Calculations and Cross-Validation Results}

The complete cross-validation results are saved with the run records. The two values reported here are individual results. They do not represent the mean, minimum, or maximum across all temperatures.

The two appendix figures show fits that use all 16 complete DST and US06 traces for parameter identification and evaluation. Their mean RMSE across conditions is 4.86\,mV. Figure~\ref{fig:results}c separately shows two traces at temperatures excluded from parameter identification. Every RMSE uses all aligned samples in its condition.

\begin{figure}[t]
\centering
\includegraphics[width=\linewidth]{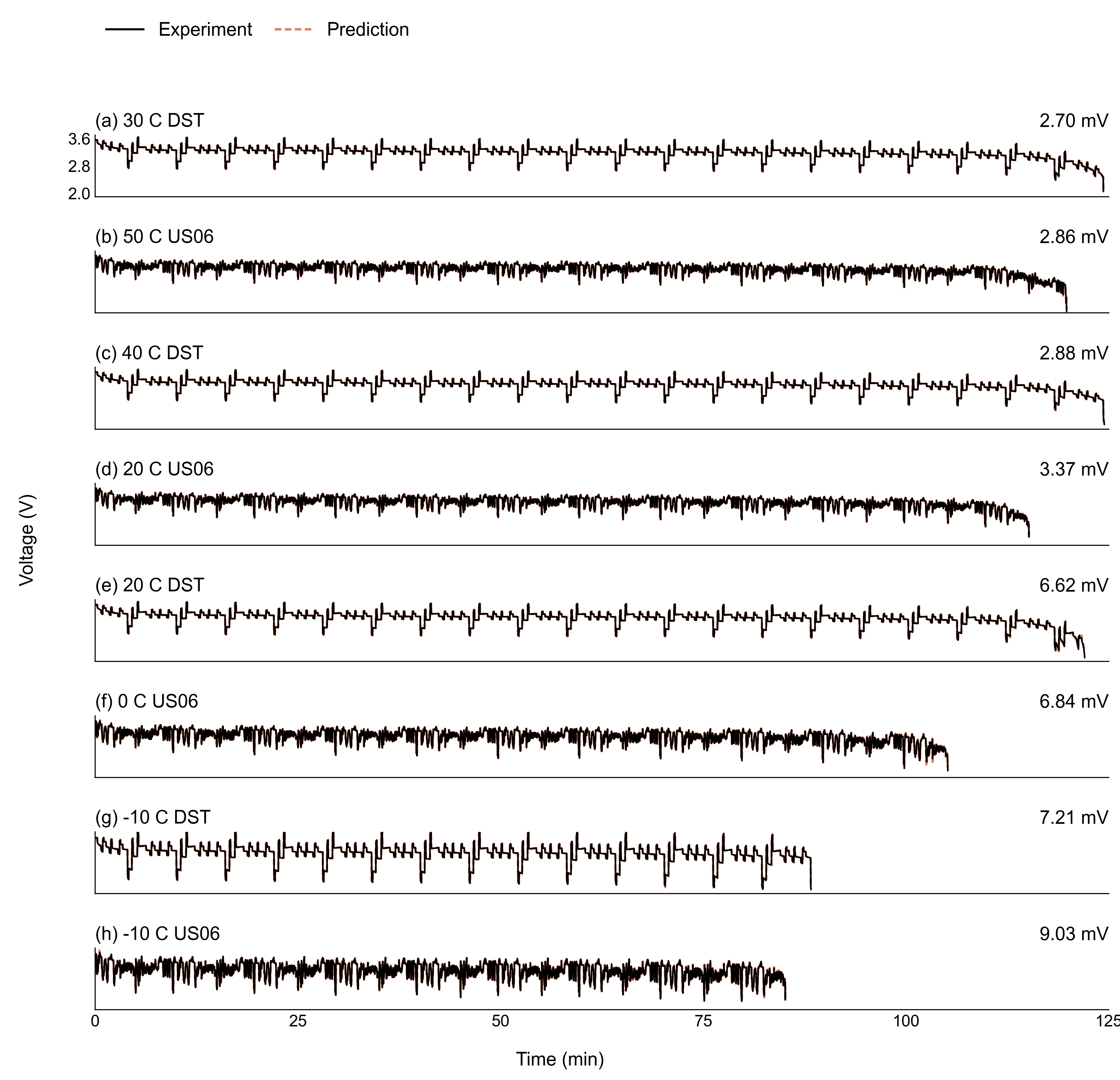}
\caption{Selected complete-trace runs of the physics-based voltage model. Figure~\ref{fig:results}c reports two separate held-out results.}
\label{fig:appendix-voltage-fullfit}
\end{figure}

\begin{figure}[t]
\centering
\includegraphics[width=\linewidth]{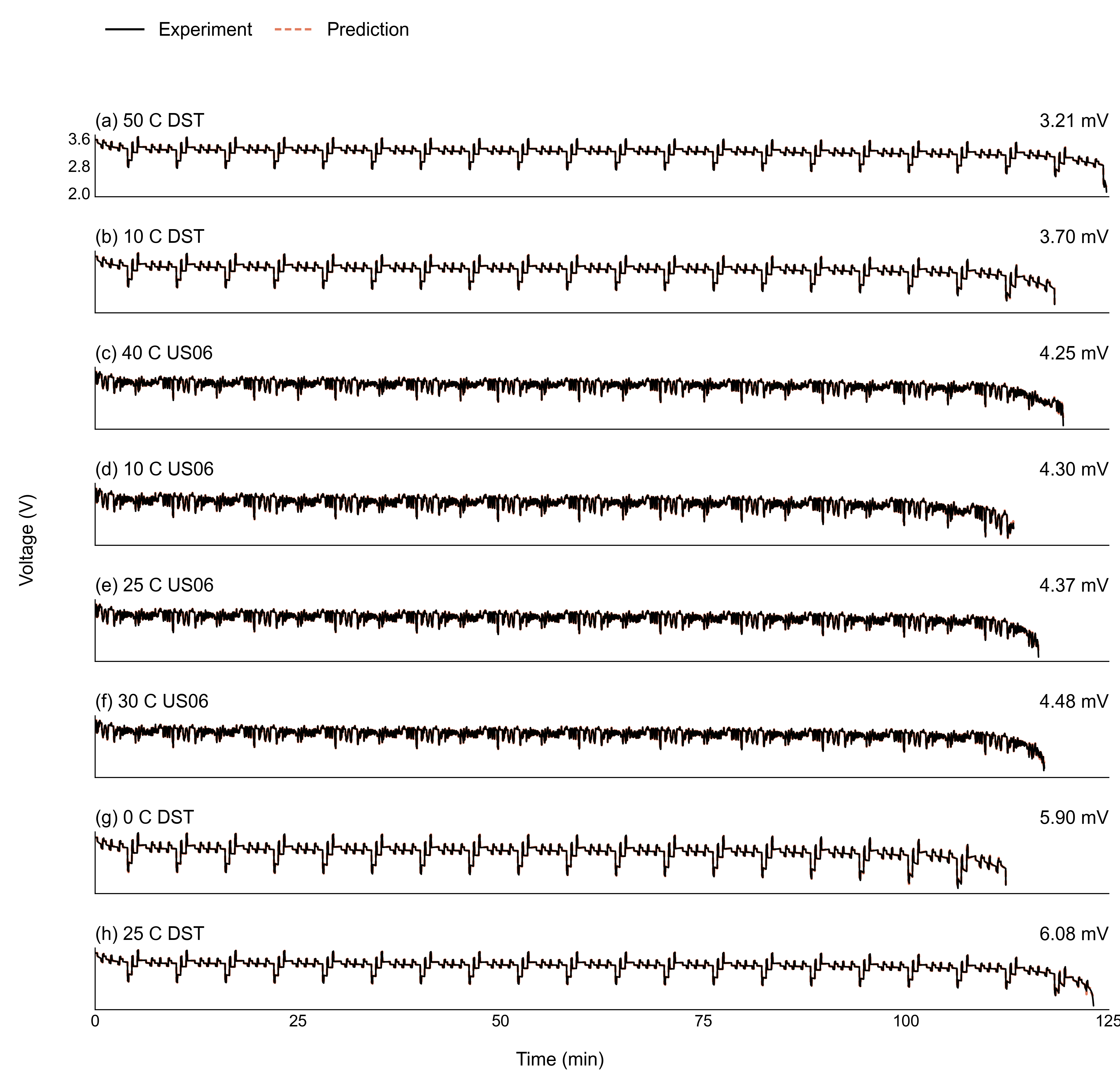}
\caption{Remaining complete-trace fits using all available data. Together with Figure~\ref{fig:appendix-voltage-fullfit}, these panels contain all 16 traces.}
\label{fig:appendix-voltage-fullfit-remaining}
\end{figure}

The appendix panels report complete model runs. The main-text panels report two results for individual runs from leave-one-temperature cross-validation of the physics-based voltage model.

\subsubsection{Reference Models}

The first reference block fixes the cell, measurement file, temperature, load protocol, physical inputs, prediction horizon, and error definition. It uses the A123 A1-007 25$^\circ$C DST trace from the public CALCE package. Both calculations receive the complete applied-current and measured-temperature sequence and report terminal-voltage RMSE over the complete trace. Each retains the calibration procedure of its own study. The values therefore provide same-condition numerical context. The physics-based run gives a 6.08\,mV result, and the published CPG-SPMT result is 32.10\,mV. The 6.08\,mV value is not held out.

\subsubsection{Comparison with Published Voltage Models}

Table~\ref{tab:appendix-voltage-reference} reports models evaluated on DST or US06 dynamic conditions with terminal-voltage RMSE in mV. The first block uses the same cell and measurement condition while retaining each study's original calibration procedure. The second block lists two EvoSim cross-validation results and reported literature values under DST or US06 conditions. This organization states the role of each value directly.

\begin{table}[!htbp]
\centering
\caption{Reported dynamic-voltage RMSE under DST or US06 operating conditions. The first block uses the same A123 A1-007 25$^\circ$C DST measurement trace and inputs.}
\label{tab:appendix-voltage-reference}
\scriptsize
\setlength{\tabcolsep}{2.5pt}
\begin{tabularx}{\linewidth}{@{}>{\raggedright\arraybackslash}p{0.37\linewidth}>{\raggedright\arraybackslash}X>{\centering\arraybackslash}p{0.18\linewidth}@{}}
\toprule
Model structure & Comparison condition & RMSE (mV) \\
\midrule
\multicolumn{3}{@{}l}{\textit{Same A123 trace and inputs; original identification procedures}} \\
Physics-based model (this work) & Full-data run; not held out & 6.08 \\
CPG-SPMT \citep{Guo2026CPGSPMT} & Published result on the same condition & 32.10 \\
\midrule
\multicolumn{3}{@{}l}{\textit{DST/US06 dynamic-voltage results reported by each study}} \\
Physics-based model (this work) & Held-out results shown in Figure~\ref{fig:results} & 7.62, 8.97 \\
NDCTNet \citep{Tu2024RemainingDischargeEnergy} & Reported results & 12.25, 13.01 \\
ESP \citep{He2024ExtendedSPM} & Reported result & 13.90 \\
PIROL \citep{Rifat2026PIROL} & Reported result & 17.80 \\
2RC-KiBaM \citep{Kasper2025KiBaM} & Reported result & 18.60 \\
2RC ECM \citep{ValderramaVelez2026LowCostValidation} & Reported result & 23.24 \\
Cui et al. \citep{Cui2023LongSequenceVoltage} & Reported results & 34.40, 49.50 \\
\bottomrule
\end{tabularx}
\end{table}

The first block shares the cell, measurement trace, current and temperature inputs, prediction horizon, and voltage metric. The calibration procedures remain those of the original studies. The second block places the two designated results from this work alongside reported DST or US06 results from other studies.

\clearpage
\section{Scientific Evolution Comparisons}
\label{sec:appendix-scientific-learning}

This section evaluates Scientific Evolution at three levels. A controlled comparison tests whether the module changes the next model decision. A sequence of four investigations traces repeated evidence, procedure update, later use, and further revision. Three related-task records and 45 paired experiments test whether the resulting procedures improve later work.

Scientific Evolution converts an earlier investigation into a checked procedure. A procedure contains an action, its expected effect, its checks, and the conditions supported by the available evidence. These parts define one usable research method. The controlled comparison therefore evaluates the module as one operation. Baseline receives no earlier record. Naive Memory receives a chronological summary produced by EvoSim. Scientific Evolution receives the procedure that EvoSim produced and checked from the same investigation. Both records are completed before the comparison runs begin. The comparison measures how Scientific Evolution changes the next model decision.

Table~\ref{tab:appendix-evolution-evidence-map} states the question answered by each experiment. The four-step record establishes recursive updating. A model result changes a procedure. The revised procedure changes the next investigation. That investigation then supplies evidence for another update. The three-condition comparison measures the resulting change in the next model decision. The related-task records show use beyond the source investigation. The 45 paired experiments measure the effect on repeated later tasks.

\begin{table}[t]
\centering
\caption{Evidence used to evaluate Scientific Evolution.}
\label{tab:appendix-evolution-evidence-map}
\scriptsize
\setlength{\tabcolsep}{3pt}
\renewcommand{\arraystretch}{1.08}
\begin{tabularx}{\linewidth}{@{}>{\raggedright\arraybackslash}p{0.22\linewidth}>{\raggedright\arraybackslash}p{0.39\linewidth}>{\raggedright\arraybackslash}X@{}}
\toprule
Question & Evidence & Conclusion supported \\
\midrule
Does Scientific Evolution produce a usable procedure? & An earlier continuous-state model gives 88.518\,mV error against the reference implementation. EvoSim diagnoses the discrete state update, checks the corrected implementation at 4.1136\,mV, and saves the procedure. & EvoSim converts its own model evidence into a checked action with stated conditions. \\
Does the module improve the next decision? & Baseline, Naive Memory, and Scientific Evolution use the same comparison task, starting model, data, solver, checks, and maximum of 12 attempts. The first three attempts are the matched comparison. Naive Memory and Scientific Evolution use the same source investigation. & Scientific Evolution reaches 22.015\,mV. The chronological summary remains at 89.383\,mV. \\
Does recursive updating occur? & Four successive research steps connect failure diagnosis, a checked correction, a 24-condition application, and the controlled comparison. & Each of the first three results changes the procedure used in the following investigation. The following result supplies evidence for the next update. \\
Does the updated procedure help later work? & The procedure changes the later state implementation and reaches a valid lower-error model on attempt 2. Three related-task records report fewer revisions or lower error after later use. & The updates change later model decisions, implementation effort, and numerical results within and across tasks. \\
Does the effect persist across repeated tasks? & Forty-five paired experiments compare Baseline with Scientific Evolution after the procedures and pair list are fixed. & Model / physics failures decrease from 33 to 21; the exact paired test remains significant after correction across six labels. \\
\bottomrule
\end{tabularx}
\end{table}

\subsection{Controlled Test of the Scientific Evolution Module}
\label{sec:appendix-memory-ablation}

This comparison fixes the task, starting model, data, solver, tools, model checks, and maximum of 12 attempts. The first three attempts are the matched comparison. The remaining attempts are a separate check of continued search. Before the comparison runs, EvoSim completes an earlier investigation and records its model changes, failures, and validation results. Section~\ref{sec:appendix-capability-trace} follows how this record is created and revised. Naive Memory receives a chronological summary of the investigation. Scientific Evolution receives the checked procedure produced from the same evidence. Baseline receives no earlier record. All three conditions restart from the same model. Both records are completed before the comparison runs begin. The comparison measures whether the procedure changes the later modeling decisions.

\paragraph{Task and metric.}
The task provides current, temperature, and measured voltage for a 30$^\circ$C DST profile. The starting model is a four-state CPG-SPMT model. The task asks the system to repair the state evolution and its connection to the physics-based model voltage calculation. A candidate must pass the validity rules below. The reported metric is the error over $N=7{,}461$ experimental voltage samples,
\begin{equation}
\operatorname{RMSE}_{\mathrm{exp}}
=1000\sqrt{\frac{1}{N}\sum_{k=1}^{N}
\left(V^{\mathrm{phys}}_k-V^{\mathrm{exp}}_k\right)^2}.
\end{equation}
Here $V^{\mathrm{phys}}_k$ is the voltage produced by the submitted physics-based model and $V^{\mathrm{exp}}_k$ is the measured voltage at sample $k$. Both are in V. The factor 1000 reports the RMSE in mV.

\paragraph{Conditions.}
All three runs begin with the same four-state model and fixed task inputs. Baseline receives no record from the earlier investigation. Naive Memory receives EvoSim's chronological summary of that investigation. Scientific Evolution receives the procedure that EvoSim derived and checked from the same investigation. This procedure states the action, the conditions under which it applies, and the checks it must pass. Each run builds candidates from the common starting model and applies the same checks. The primary comparison uses the first three attempts from each condition. Scientific Evolution reaches its best valid model on attempt 2 and completes a third diagnostic attempt. Baseline and Naive Memory continue to 12 attempts in a separate persistence check.

\begin{table}[t]
\centering
\caption{EvoSim-generated records available to the two conditions. Both records come from the same earlier investigation.}
\label{tab:appendix-record-input-content}
\scriptsize
\setlength{\tabcolsep}{3pt}
\renewcommand{\arraystretch}{1.08}
\begin{tabularx}{\linewidth}{@{}>{\raggedright\arraybackslash}p{0.19\linewidth}>{\raggedright\arraybackslash}p{0.38\linewidth}>{\raggedright\arraybackslash}X@{}}
\toprule
Condition & Earlier record & Information available for the next decision \\
\midrule
Naive Memory & ``A later version represented the sample-wise state recurrence explicitly and reduced the 30 degC DST COMSOL-to-Python voltage RMSE to about 4 mV.'' & Identifies the general cause and records the earlier result. It does not state the recurrence, sample timing, state binding, initial-state check, temperature rule, or invalid-output conditions. \\
Scientific Evolution & ``For every $k>0$, use current and temperature from sample $k$'' and ``Export time, current, temperature, all four COMSOL dependent states, and terminal voltage directly from COMSOL.'' & Gives the exact recurrence, initial values, one temperature conversion, four dependent-state equations, direct-output checks, supported conditions, and failure conditions. \\
\bottomrule
\end{tabularx}
\end{table}

EvoSim produces both inputs from the same earlier investigation before the comparison runs begin. Naive Memory preserves the chronological account. Scientific Evolution records a specific action, the conditions for its use, and the checks it passed. The controlled comparison evaluates these two records under the same task conditions.

\paragraph{Candidate validity.}
A structurally valid model exports a complete trajectory with finite values. The four required states must be dependent variables of the multiphysics model. The model must use the fixed terminal-voltage relation. Time, input, state, and voltage must come directly from the multiphysics solver. The same checks apply in all three conditions. A zero-order-hold (ZOH) update keeps each input fixed over one sample interval. The retained procedure uses the exact ZOH solution for the state update. Calculating the states and voltage outside the multiphysics solver fails the implementation check even when it gives a lower numerical error.

\begin{table}[!htbp]
\centering
\caption{Best valid result produced within the first three attempts (physics-based model--experimental-voltage RMSE).}
\label{tab:appendix-memory-inputs}
\small
\renewcommand{\arraystretch}{1.08}
\begin{tabularx}{\linewidth}{@{}>{\raggedright\arraybackslash}p{0.20\linewidth}>{\raggedright\arraybackslash}X>{\centering\arraybackslash}p{0.20\linewidth}@{}}
\toprule
Condition & Earlier record available to the run & \shortstack{Experimental-voltage\\RMSE (mV)} \\
\midrule
Baseline & No retained record from the earlier investigation & 89.478\,mV \\
Naive Memory & EvoSim-generated chronological summary of the same earlier investigation & 89.383\,mV \\
Scientific Evolution & EvoSim-generated procedure produced and checked from the same earlier investigation & \textbf{22.015\,mV} \\
\bottomrule
\end{tabularx}
\end{table}

The Naive Memory and Scientific Evolution records come from the same earlier investigation. Naive Memory preserves the chronological account of attempts and results. Scientific Evolution preserves the action, the conditions under which it applies, and the checks that it passed. The chronological summary remains near the Baseline result. The checked procedure changes the sequence of model changes and reaches the lower-error valid model.

Let $B$ denote the cumulative number of submitted candidates. Table~\ref{tab:appendix-memory-budget} separates the common $B=3$ comparison from the later Baseline and Naive Memory attempts. Baseline and Naive Memory use additional submissions without changing their best valid result. Scientific Evolution reaches its lower-error valid model at $B=2$. The Scientific Evolution run ends after its third submission, so the later columns contain no value for that condition.

\begin{table}[!htbp]
\centering
\caption{Best valid RMSE in the first three submissions and during the Baseline and Naive Memory persistence check.}
\label{tab:appendix-memory-budget}
\small
\setlength{\tabcolsep}{2pt}
\renewcommand{\arraystretch}{1.08}
\begin{tabularx}{\linewidth}{@{}>{\centering\arraybackslash}p{0.14\linewidth}>{\centering\arraybackslash}p{0.16\linewidth}>{\centering\arraybackslash}p{0.16\linewidth}>{\centering\arraybackslash}p{0.19\linewidth}>{\raggedright\arraybackslash}X@{}}
\toprule
Budget level & Baseline & Naive Memory & Scientific Evolution & Interpretation \\
 & \multicolumn{3}{c}{RMSE (mV)} & \\
\midrule
$B=3$ & 89.478 & 89.383 & \textbf{22.015} & First three submissions \\
$B=6$ & 89.478 & 89.383 & -- & Baseline and Naive Memory persistence check \\
$B=12$ & 89.478 & 89.383 & -- & Baseline and Naive Memory persistence check \\
\bottomrule
\end{tabularx}

\vspace{2pt}
\begin{minipage}{0.96\linewidth}
\footnotesize A dash indicates that no Scientific Evolution submission was executed at that budget level. Values are not carried forward as new observations.
\end{minipage}
\end{table}

Each condition is one autonomous agent run. The primary comparison uses the first three submissions from every condition. The $B=6$ and $B=12$ entries continue the same Baseline and Naive Memory runs. They measure persistence within those runs. Section~\ref{sec:appendix-paired-analysis} reports 45 separate pairs across repeated runs and tasks.

The procedure produced by the earlier EvoSim investigation contains the following exact zero-order-hold (ZOH) state update. The sampling interval is $\Delta t=1$\,s. For each electrode, the update first defines
\begin{equation}
\begin{aligned}
u_{n,k}&=I_k\Delta t/b_n, & u_{p,k}&=-I_k\Delta t/b_p,\\
\alpha_{n,k}&=30\Delta t/a_{n,k}, & \alpha_{p,k}&=30\Delta t/a_{p,k},\\
\beta_{n,k}&=19I_k\Delta t/(7b_n), & \beta_{p,k}&=-19I_k\Delta t/(7b_p).
\end{aligned}
\label{eq:appendix-cpg-zoh-parameters}
\end{equation}
Here $I_k$ is the applied current in A at sample $k$. Positive current denotes charge. The index $e\in\{n,p\}$ denotes the negative or positive electrode. The parameter $b_e$ is an electrode capacity scale in C. The temperature-dependent transport time $a_{e,k}$ is in s. Thus, $u_{e,k}$, $\alpha_{e,k}$, and $\beta_{e,k}$ are dimensionless. The quantity $u_{e,k}$ is the change in average filling fraction during sample $k$. The quantity $\alpha_{e,k}$ is the decay rate multiplied by the sample duration. The quantity $\beta_{e,k}$ is the current forcing over the same sample. The constants 30 and $19/7$ come from the reduced particle-diffusion equations.
For $e\in\{n,p\}$ and $k>0$, the ZOH recurrence is
\begin{equation}
\begin{aligned}
\bar{x}_{e,k}&=\bar{x}_{e,k-1}+u_{e,k},\\
\delta_{e,k-1}&=x^{\mathrm{surf}}_{e,k-1}-\bar{x}_{e,k-1},\\
\delta_{e,k}&=\delta_{e,k-1}e^{-\alpha_{e,k}}
+\frac{\beta_{e,k}-u_{e,k}}{\alpha_{e,k}}\left(1-e^{-\alpha_{e,k}}\right),\\
x^{\mathrm{surf}}_{e,k}&=\bar{x}_{e,k}+\delta_{e,k}.
\end{aligned}
\label{eq:appendix-cpg-zoh-update}
\end{equation}
The average filling fraction is $\bar{x}_{e,k}$. The surface filling fraction is $x^{\mathrm{surf}}_{e,k}$. Their difference is $\delta_{e,k}$. The $k=0$ sample retains the prescribed initial states. Each later update uses current and temperature from sample $k$. Temperature is converted from Celsius to Kelvin once. The four state trajectories are assigned to four algebraic dependent states of the physics-based model. The multiphysics solver then evaluates the fixed surface outputs
\begin{equation}
\begin{aligned}
C_{n,k}&=\operatorname{clip}_{[\epsilon_x,1-\epsilon_x]}
\!\left(x^{\mathrm{surf}}_{n,k}+\frac{a_{n,k}}{105b_n}I_k\right),\\
C_{p,k}&=\operatorname{clip}_{[\epsilon_x,1-\epsilon_x]}
\!\left(x^{\mathrm{surf}}_{p,k}-\frac{a_{p,k}}{105b_p}I_k\right).
\end{aligned}
\label{eq:appendix-cpg-zoh-surface}
\end{equation}
Here $C_{n,k}$ and $C_{p,k}$ are the bounded surface filling fractions supplied to the equilibrium-potential and charge-transfer equations. Equation~\ref{eq:appendix-voltage-state-bound} defines the bound and $\epsilon_x$. The solver then evaluates the terminal-voltage equation given above.

\paragraph{Modeling attempts.}
Tables~\ref{tab:appendix-memory-slots} and \ref{tab:appendix-memory-continuation} report every attempted model change. Values in parentheses are experimental-voltage RMSE in mV. A candidate is marked failed when it produces no complete finite trajectory. It is marked invalid when it produces numerical output but fails a model check.

\begin{table}[!htbp]
\centering
\caption{First three modeling attempts in the three-condition comparison.}
\label{tab:appendix-memory-slots}
\scriptsize
\setlength{\tabcolsep}{3pt}
\renewcommand{\arraystretch}{1.06}
\begin{tabularx}{\linewidth}{@{}rXXX@{}}
\toprule
Attempt & Baseline & Naive Memory & Scientific Evolution \\
\midrule
1 & Continuous physics-based model state equations (89.478; valid) & Continuous Global Equations starting model (89.383; valid) & Setup failed; no model output \\
2 & Previous-sample current and temperature (89.666; valid) & Discrete states and event update; build failure & Exact ZOH, one temperature conversion, and four algebraic states (22.015; valid) \\
3 & Exact ZOH interpolation without the required dependent-state representation (83.638; invalid) & Event reinitialization leaves states at their initial values (394.480; valid) & Nearest/sample-hold state transfer; diagnostic export, not retained \\
\bottomrule
\end{tabularx}
\end{table}

\begin{table}[!htbp]
\centering
\caption{Baseline and Naive Memory attempts after the first three attempts. A candidate is invalid when it bypasses the required solver-based state and voltage equations.}
\label{tab:appendix-memory-continuation}
\scriptsize
\setlength{\tabcolsep}{4pt}
\renewcommand{\arraystretch}{1.06}
\begin{tabularx}{\linewidth}{@{}rXX@{}}
\toprule
Attempt & Baseline & Naive Memory \\
\midrule
4 & Solver bypass: bypasses the required model equations and returns a supplied target output (invalid) & Global event selection activates updates, but the surface state collapses to the average state (89.578; valid) \\
5 & Repeats the same solver bypass (invalid) & Explicit-Euler surface-state update diverges; failed \\
6 & Repeats the same solver bypass (invalid) & Analytic relaxation restores stability (89.578; valid) \\
7 & Repeats the same solver bypass (invalid) & Exact matrix recurrence, average state first (89.757; valid) \\
8 & Repeats the same solver bypass (invalid) & Surface-state update created first (89.757; valid) \\
9 & Repeats the same solver bypass (invalid) & Shifted zero-order-hold current (89.573; valid) \\
10 & Repeats the same solver bypass (invalid) & Shifted temperature-dependent parameters (89.573; valid) \\
11 & Repeats the same solver bypass (invalid) & Direct output replaces a required state output (89.573; invalid) \\
12 & Repeats the same solver bypass (invalid) & Repeated direct-output substitution (89.573; invalid) \\
\bottomrule
\end{tabularx}
\end{table}

The attempt log shows the sequence taken by each condition. Baseline identifies a discrete-state approach at attempt 3. It then returns a supplied voltage instead of calculating voltage from the required model states. This is a solver bypass. The implementation check rejects the submission because the required dependent states and terminal-voltage equation do not produce the reported output. Baseline repeats the same approach through attempt 12. Naive Memory continues with event-based state updates. Several stable variants collapse the surface state toward the average state. Later output variants fail the implementation check. Scientific Evolution changes the state representation and checks the state update, temperature conversion, dependent variables, and voltage output together. Its second attempt is the first valid low-error solver-based model.

\subsection{Use After Scientific Evolution}
\label{sec:appendix-four-comparisons}

The controlled voltage task shows that the module changes the later model implementation. Three additional records show how procedures produced during an earlier investigation are used in a related task. Figure~\ref{fig:evolution} summarizes these outcomes. In each comparison, the task instruction, starting model, data, simulator, tools, and model checks remain fixed. The saved procedure changes the order or content of the later model decisions.

\begin{table}[t]
\centering
\caption{Outcomes of the four task-level model-development comparisons.}
\label{tab:appendix-ablation-controls}
\scriptsize
\setlength{\tabcolsep}{3pt}
\renewcommand{\arraystretch}{1.08}
\begin{tabularx}{\linewidth}{@{}>{\raggedright\arraybackslash}p{0.17\linewidth}>{\raggedright\arraybackslash}p{0.28\linewidth}>{\raggedright\arraybackslash}p{0.23\linewidth}>{\raggedright\arraybackslash}X@{}}
\toprule
Task & Fixed task and acceptance check & Observed outcome & What the outcome demonstrates \\
\midrule
Voltage-model implementation & Four required physics-based model states, fixed terminal-voltage relation, complete finite export & Baseline 89.478\,mV; Naive Memory 89.383\,mV; Scientific Evolution 22.015\,mV & The checked procedure changes the implementation and lowers the voltage error \\
Capacity matching & Specified cell configurations; every candidate must build, solve, and export & Required attempts decrease from 8 to 2 & A retained procedure reduces repeated model revisions \\
P2D--thermal coupling & Coupled model must solve and export the complete temperature trace & Model changes decrease from 10 to 1 & A retained procedure changes how the coupling investigation is organized \\
Thermal-mechanism selection & Particle size, polarization, boundary heat, and convection use one temperature metric & Temperature RMSE decreases from 9.30 to 3.74$^\circ$C & A common-check comparison changes both candidate selection and prediction error \\
\bottomrule
\end{tabularx}
\end{table}

\begin{table}[t]
\centering
\caption{Use of saved procedures in three related modeling tasks. Each procedure is completed before the later task begins. Each later task uses its own acceptance check.}
\label{tab:appendix-cross-task-reuse}
\scriptsize
\setlength{\tabcolsep}{3pt}
\renewcommand{\arraystretch}{1.08}
\begin{tabularx}{\linewidth}{@{}>{\raggedright\arraybackslash}p{0.18\linewidth}>{\raggedright\arraybackslash}p{0.22\linewidth}>{\raggedright\arraybackslash}p{0.25\linewidth}>{\raggedright\arraybackslash}X@{}}
\toprule
Earlier investigation & Later investigation & Reused modeling decision & Result in the later task \\
\midrule
Capacity investigation & Specified-configuration capacity-matching task & Apply the tested capacity-matching order and stop only after the configuration, solve, and export checks agree & Required attempts decrease from 8 to 2 \\
Thermal-mechanism investigation & Subsequent P2D--thermal coupling task & Compare independent coupling mechanisms before repeatedly modifying one candidate; retain only a candidate that passes the common temperature check & Sequential model changes decrease from 10 to 1 \\
Thermal-mechanism investigation & Thermal-mechanism selection target & Use one temperature metric and compare particle-size, polarization, boundary-heat, and convection candidates under the same starting conditions & Temperature RMSE decreases from 9.30 to 3.74$^\circ$C \\
\bottomrule
\end{tabularx}
\end{table}

Each row answers a practical question. It shows whether the implementation is correct, how many revisions are needed, how a coupled model is organized, or how candidate physics are selected under a common check. In every row, the earlier investigation produces the procedure first. The later task then applies it to its own model and data. The result is measured with that task's acceptance check.

For the P2D--thermal row, electrochemical reaction and transport losses generate heat in the cell thermal equation. The solved temperature then changes the temperature-dependent electrochemical parameters. These two directions define the coupled model. Supplying measured temperature or applying a temperature correction after the solve gives a temperature-driven voltage model.

The four outcomes cover implementation accuracy, revision effort, multiphysics coupling, and mechanism selection. Together they show how an earlier research record changes the modeling approach and the resulting decision. The voltage case shows reuse within one task. The other three cases show reuse in related tasks. Each target task uses its own metric and acceptance check.

A procedure reused across tasks contains three parts: the observation that starts the investigation, the action taken, and the checks for that action. The target task supplies its own model, data, parameters, and prediction metric. Fitted values and physical conclusions remain specific to the task that produced them. A procedure is reused only when the target task requires the same action and can repeat the same checks. For example, the P2D--thermal task reuses the procedure for comparing several mechanisms under one metric. It evaluates its own heat-generation and heat-transfer mechanisms.

\subsection{How Scientific Evolution Produces a Procedure}
\label{sec:appendix-capability-trace}

The following record shows how Scientific Evolution produces the procedure used in the controlled comparison. EvoSim first investigates the four-state voltage model. It rejects the continuous state implementation. It then extracts an exact discrete update from the diagnosed failure. EvoSim checks the update and records its supported conditions before later use.

The earlier investigation, its review, and the decision to retain the procedure are stored together. These records show how EvoSim derives a procedure from a failed model attempt and applies it in a later run.

The procedure is revised when a later result changes its supported conditions or checks. Table~\ref{tab:appendix-recursive-procedure-updates} follows four successive investigations. A failed model produces the diagnosis used in the correction. The corrected model produces a checked procedure used across 24 conditions. Those results expand its supported conditions. The controlled comparison adds direct evidence for the temperature and state-output checks. In each of the first three rows, the recorded result changes the procedure used in the next row.

\begin{table}[t]
\centering
\caption{Successive evidence and updates for the exact zero-order-hold procedure.}
\label{tab:appendix-recursive-procedure-updates}
\scriptsize
\setlength{\tabcolsep}{3pt}
\renewcommand{\arraystretch}{1.08}
\begin{tabularx}{\linewidth}{@{}>{\raggedright\arraybackslash}p{0.16\linewidth}>{\raggedright\arraybackslash}p{0.29\linewidth}>{\raggedright\arraybackslash}p{0.29\linewidth}>{\raggedright\arraybackslash}X@{}}
\toprule
Research step & New evidence & Change to the saved procedure & Observed later use \\
\midrule
Failed earlier model & Continuous state equations solve but give 88.518\,mV RMSE against the official discrete reference implementation. The surface states contain the largest differences. & Record the discrete state update as the cause to test before changing parameters or the voltage equation. & The next model implements the exact sample-wise zero-order-hold update. \\
Single-condition correction & The corrected 30$^\circ$C DST model gives 4.1136\,mV RMSE against the reference and passes the state and output checks. & Retain the state update, temperature conversion, four-state representation, and output checks as a reusable procedure. & A later investigation applies the procedure to every available temperature and drive-cycle condition. \\
Twenty-four-condition application & All 24 conditions solve. Their mean RMSE against the reference implementation is 4.4625\,mV. & Expand the supporting evidence from one trace to 24 conditions while retaining the one-second CPG-SPMT equations and input rules as the stated scope. & The checked procedure becomes available to the later controlled voltage-modeling task. \\
Controlled comparison & The first application detects a repeated temperature conversion. The next attempt corrects the conversion, represents all four dependent states, and gives 22.015\,mV RMSE against measured voltage. & Add the failed double conversion as direct evidence for the temperature check and record the state-output check with the valid implementation. & The procedure changes the modeling sequence and reaches the valid low-error model on attempt 2. \\
\bottomrule
\end{tabularx}
\end{table}

\begin{table}[t]
\centering
\caption{Model and procedure decisions in the fixed four-state voltage task.}
\label{tab:appendix-capability-trace}
\scriptsize
\setlength{\tabcolsep}{3pt}
\renewcommand{\arraystretch}{1.08}
\begin{tabularx}{\linewidth}{@{}>{\raggedright\arraybackslash}p{0.18\linewidth}>{\raggedright\arraybackslash}p{0.50\linewidth}>{\raggedright\arraybackslash}X@{}}
\toprule
Record element & Recorded evidence & Consequence for the next decision \\
\midrule
Earlier candidate & The earlier voltage investigation used continuous state equations in the physics-based model. The candidate passed the finite-output and input-value checks but failed the target state and voltage comparison. & The discrepancy motivates separate decisions about the model and the state-update procedure. \\
Model decision & Reject the continuous implementation for the stated task. & Use the diagnosed discrepancy to examine the state-update procedure. \\
Proposed procedure & The logs support an exact sample-wise ZOH update. Convert the trace temperature once. Use nearest-neighbor current and temperature inputs with the recorded half-step timing. Represent the computed states as linear time functions on the one-second grid. Preserve the stated equations, voltage bounds, and checked solver settings. & Define a specific operation and the checks needed to evaluate it. \\
Procedure decision & Retain the exact ZOH procedure for the one-second, 30$^\circ$C DST task after the implementation checks pass. & Make the procedure available with the same unit, state-variable, and output checks. \\
Saved procedure & EvoSim stores the procedure, its source evidence, supported conditions, and file list with the run. & A later attempt can retrieve the operation and its conditions from one record. \\
First later use & The first application converts a $^\circ$C value to Kelvin twice and fails the submission check. The next attempt uses one conversion, four algebraic states in the physics-based model, and the fixed output format. & Correct the temperature conversion before testing the retained operation. \\
Later result & The second target attempt exports 7,461 finite rows, passes the implementation checks, and gives 22.015\,mV physics-based model--experimental-voltage RMSE. & The record changes the model sequence and yields the valid lower-error result under the stated conditions. \\
\bottomrule
\end{tabularx}
\end{table}

The four rows show three consecutive updates. The failed model changes the next implementation test. The corrected implementation creates the procedure used across 24 conditions. The 24-condition result expands the supported conditions before the controlled comparison. Each result changes the saved procedure, and that procedure changes the following investigation.

\subsection{Evaluation on Later Tasks After Scientific Evolution}
\label{sec:appendix-paired-analysis}

This experiment evaluates the effect of Scientific Evolution on later research. Scientific Evolution produces and checks the procedures before the paired runs begin. The procedures are then fixed during evaluation. Each pair measures how the retained procedures affect a later run.

The successive-update record in Table~\ref{tab:appendix-recursive-procedure-updates} and the paired experiment answer different questions. The record shows how new evidence changes a saved procedure. The paired experiment measures the effect of the saved procedures on later tasks. The evaluation results do not change the procedures being evaluated.

The analysis covers 45 paired experiments across 11 modeling tasks. Each pair contains one Baseline run and one Scientific Evolution run for the same task. The two runs use the same starting model, data, solver, tools, model checks, evaluation rules, and budget. The analysis contains 90 distinct runs, with each run assigned to one pair. The available procedures, model-selection rules, and pair list are fixed before evaluation. The pair is the unit of analysis.

\begin{table}[t]
\centering
\caption{Run accounting for the paired-run analysis.}
\label{tab:appendix-paired-accounting}
\small
\setlength{\tabcolsep}{3pt}
\begin{tabularx}{\linewidth}{@{}>{\raggedright\arraybackslash}p{0.72\linewidth}>{\centering\arraybackslash}X@{}}
\toprule
Quantity & Count \\
\midrule
Modeling tasks with repeated runs & 11 \\
Baseline runs & 45 \\
Scientific Evolution runs & 45 \\
Unique runs & 90 \\
Baseline--Scientific Evolution pairs & 45 \\
\bottomrule
\end{tabularx}
\end{table}

Each pair compares two runs from the same task. The two runs share the fixed task inputs and evaluation rules. Their experimental condition is fixed before execution.

The six failure labels are assigned manually after Scientific Validation. The labels use the saved evidence on implementation, solution, and agreement with measurements. The same written definitions are used for every task and condition. One reviewer assigns the initial labels. A second reviewer checks the final report, validation output, solver record, and model files. The reviewers resolve disagreements from these saved files before counting the labels. A failure is counted when it remains unresolved or materially affects the final result. A run can receive more than one label. Repeated instances of one failure within a run count once. The labels describe the saved run and do not alter its model-acceptance result. In the next table, ``Baseline-only'' means that a label occurs only in the Baseline run of a pair. ``Both'' means that it occurs in both runs. ``Evolution-only'' means that it occurs only in the Scientific Evolution run.

\begin{table}[t]
\centering
\caption{Operational definitions used for failure-mode labels.}
\label{tab:appendix-failure-definitions}
\small
\begin{tabularx}{\linewidth}{@{}>{\raggedright\arraybackslash}p{0.24\linewidth}>{\raggedright\arraybackslash}X@{}}
\toprule
Failure mode & Recorded condition \\
\midrule
False completion & Completion is declared while a required output, check, or scientific result still fails \\
Tool operation & Model inspection, build, save, export, or data access is executed incorrectly \\
Model / physics & A required physical process is absent or inactive, or the tested model form cannot reproduce the target behavior under the task criterion \\
Data / protocol & Inputs, units, initial conditions, stopping rules, time alignment, or data roles are bound incorrectly \\
Numerical solver & The solve fails, diverges, becomes singular, or produces non-finite values \\
Looping / stagnation & Repeated actions or failed recovery do not advance the task to a supported decision \\
\bottomrule
\end{tabularx}
\end{table}

\begin{table}[t]
\centering
\caption[Failure-mode changes in 45 pairs]{Failure-mode transitions across 45 same-task pairs.}
\label{tab:appendix-failure-transitions}
\scriptsize
\setlength{\tabcolsep}{2pt}
\renewcommand{\arraystretch}{1.08}
\begin{tabularx}{\linewidth}{@{}>{\raggedright\arraybackslash}p{0.22\linewidth}>{\centering\arraybackslash}X>{\centering\arraybackslash}X>{\centering\arraybackslash}X>{\centering\arraybackslash}X>{\centering\arraybackslash}X>{\centering\arraybackslash}X@{}}
\toprule
Failure mode & \shortstack{Baseline\\count} & \shortstack{Scientific\\Evolution\\count} & \shortstack{Baseline-\\only} & Both & \shortstack{Evolution-\\only} & \shortstack{Net\\decrease (\%)} \\
\midrule
False completion & 12 & 0 & 12 & 0 & 0 & 100.0\% \\
Tool operation & 19 & 2 & 17 & 2 & 0 & 89.5\% \\
Model / physics & 33 & 21 & 15 & 18 & 3 & 36.4\% \\
Data / protocol & 14 & 0 & 14 & 0 & 0 & 100.0\% \\
Numerical solver & 5 & 2 & 5 & 0 & 2 & 60.0\% \\
Looping / stagnation & 25 & 5 & 21 & 4 & 1 & 80.0\% \\
\bottomrule
\end{tabularx}
\end{table}

Net decrease is $(\mathrm{Baseline}-\mathrm{Scientific\ Evolution})/\mathrm{Baseline}$. Across the paired experiments, Scientific Evolution shows fewer false completions, data/protocol failures, tool failures, and stagnating runs. Model and physical-mechanism failures remain the main unresolved class.

\paragraph{Statistical test.}
For each failure mode, the comparison is determined by the pairs in which the two conditions differ. Let $b$ be the number of Baseline-only pairs and let $c$ be the number of Scientific-Evolution-only pairs. Under the null hypothesis that the two conditions have the same failure probability, either direction is equally likely within a discordant pair. We therefore apply a two-sided exact McNemar test, which evaluates $b$ conditional on $b+c$ under a binomial probability of $0.5$. The test is exact and does not use a large-sample approximation. Because the same analysis is applied to six failure modes, the six $p$-values are adjusted together with the Holm procedure. Table~\ref{tab:appendix-failure-significance} reports the discordant counts, exact $p$-values, and adjusted $p$-values.

\begin{table}[t]
\centering
\caption{Exact paired tests for the failure-mode transitions.}
\label{tab:appendix-failure-significance}
\small
\setlength{\tabcolsep}{3pt}
\begin{tabularx}{\linewidth}{@{}>{\raggedright\arraybackslash}p{0.31\linewidth}>{\centering\arraybackslash}X>{\centering\arraybackslash}X>{\centering\arraybackslash}X>{\centering\arraybackslash}X@{}}
\toprule
Failure mode & \shortstack{Baseline-\\only ($b$)} & \shortstack{Evolution-\\only ($c$)} & \shortstack{Exact\\$p$-value} & \shortstack{Holm-adjusted\\$p$-value} \\
\midrule
False completion & 12 & 0 & 0.000488 & 0.001465 \\
Tool operation & 17 & 0 & 0.000015 & 0.000076 \\
Model / physics & 15 & 3 & 0.007538 & 0.015076 \\
Data / protocol & 14 & 0 & 0.000122 & 0.000488 \\
Numerical solver & 5 & 2 & 0.453125 & 0.453125 \\
Looping / stagnation & 21 & 1 & 0.000011 & 0.000066 \\
\bottomrule
\end{tabularx}
\end{table}

For model / physics failures, the observed rate decreases from $33/45$ (73.3\%) to $21/45$ (46.7\%), an absolute decrease of 26.7 percentage points and a relative decrease of 36.4\%. Among the 18 discordant pairs, 15 favor Scientific Evolution and 3 favor Baseline. The matched odds ratio is therefore $15/3=5.00$, with an exact 95\% confidence interval of 1.41--26.94. The exact McNemar result is $p=0.0075$, and it remains significant after correction across all six failure modes ($p_{\mathrm{Holm}}=0.0151$). False completion, data/protocol failure, tool-operation failure, and looping/stagnation also remain significant after correction. Numerical-solver failure does not show a significant difference. These results show that the reductions are consistent across matched experiments and are larger than expected from balanced pair-to-pair variation under the null hypothesis.

\begin{table}[t]
\centering
\caption{Number of tasks and runs with each failure mode. A task is counted when the failure appears in at least one run. Run counts include repeated experiments on the same task.}
\label{tab:appendix-failure-uc-equal}
\scriptsize
\setlength{\tabcolsep}{2.5pt}
\begin{tabularx}{\linewidth}{@{}>{\raggedright\arraybackslash}p{0.32\linewidth}>{\centering\arraybackslash}X>{\centering\arraybackslash}X>{\centering\arraybackslash}X>{\centering\arraybackslash}X@{}}
\toprule
Failure mode & \shortstack{Baseline\\tasks} & \shortstack{Scientific Evolution\\tasks} & \shortstack{Baseline\\runs} & \shortstack{Scientific Evolution\\runs} \\
\midrule
False completion & 1 & 0 & 12 & 0 \\
Data / protocol & 4 & 0 & 14 & 0 \\
Tool operation & 4 & 1 & 19 & 2 \\
Looping / stagnation & 6 & 2 & 25 & 5 \\
Numerical solver & 2 & 1 & 5 & 2 \\
Model / physics & 7 & 7 & 33 & 21 \\
\bottomrule
\end{tabularx}
\end{table}

The task columns show how widely each failure occurs. The run columns show how often it occurs across repeated experiments. Model / physics failures appear in 7 of the 11 tasks under both conditions. Their run count decreases from 33 under Baseline to 21 under Scientific Evolution. Of the 45 pairs, 15 contain this failure only under Baseline, 18 contain it under both conditions, and 3 contain it only under Scientific Evolution. The decrease from 33 to 21 is 36.4\%. Thus, Scientific Evolution reduces repeated model / physics failures, while the same seven tasks continue to contain difficult physical-modeling problems.

\subsubsection{Execution Efficiency}

A final-check failure is a completion attempt rejected by the final quality check. An assigned work item is one instruction sent to an execution agent. An unclosed work item is an assigned work item without a final report in the event log; its active runtime is measured to the last recorded event. A closed work item's active runtime is the time from assignment to report. Elapsed time is the time from the first to the last event in a run. The table sums each quantity over the same 45 pairs. Runs executed concurrently are counted separately, so these totals measure accumulated run time rather than calendar time.

\begin{table}[t]
\centering
\caption{Process totals and ratios across the 45 same-task pairs.}
\label{tab:appendix-efficiency}
\small
\setlength{\tabcolsep}{3pt}
\begin{tabularx}{\linewidth}{@{}>{\raggedright\arraybackslash}p{0.43\linewidth}>{\centering\arraybackslash}X>{\centering\arraybackslash}X>{\centering\arraybackslash}X@{}}
\toprule
Measure & Baseline total & Scientific Evolution total & Ratio (\%) \\
\midrule
Final-check failures & 167 & 53 & 31.7 \\
Unclosed work items & 548 & 187 & 34.1 \\
Active runtime (h) & 1086.3 & 623.6 & 57.4 \\
Elapsed time (h) & 1217.7 & 708.7 & 58.2 \\
Assigned work items & 3792 & 3169 & 83.6 \\
\bottomrule
\end{tabularx}
\end{table}

These totals show less unfinished work and less active runtime in the paired experiments. The ratio is the Scientific Evolution total divided by the Baseline total. Active runtime decreases by 42.6\%. This value covers the later paired runs. The earlier investigations that produced and checked the procedures are reported separately.

Together, the experiments support four conclusions. First, Scientific Evolution improves the next model decision relative to no earlier record and a chronological summary of the same investigation. Second, EvoSim produces the tested procedure from a failed model, a corrected implementation, and validation results. Third, later results revise the procedure before it is used again. Fourth, the updated procedures change later model implementation, reduce revisions in related tasks, and reduce repeated failures and execution effort across the paired experiments.

\section{Task Collection}
\label{sec:appendix-task-collection}

This section describes the task collection and the evidence used for each scientific conclusion. It first defines the task groups and run counts. It then identifies comparisons made after a candidate model was fixed. The final table lists the nine complete evaluation records.

The evaluation below applies a fixed model to another condition or dataset and records the resulting conclusion. Section~\ref{sec:appendix-four-comparisons} studies a different question: whether a modeling procedure learned in one task changes decisions in a later task.

\subsection{Task Inventory}

The task collection spans parameter identification, forward prediction, mechanism and multiphysics studies, and model diagnosis. Parameter identification chooses model values. Forward prediction applies a fixed model to a specified input. A mechanism study asks whether a physical process is needed and how it should be coupled. Model diagnosis records why a change failed. A task family is a group of runs that share a scientific objective, observation source, and starting model.

\begin{table}[t]
\centering
\caption{Task definitions and completion requirements in the recorded run collection.}
\label{tab:appendix-task-families}
\scriptsize
\setlength{\tabcolsep}{2.5pt}
\begin{tabularx}{\linewidth}{@{}>{\raggedright\arraybackslash}p{0.19\linewidth}>{\raggedright\arraybackslash}p{0.25\linewidth}>{\raggedright\arraybackslash}p{0.23\linewidth}>{\raggedright\arraybackslash}X@{}}
\toprule
Task family & Scientific objective & Required model output & Evidence required for completion \\
\midrule
Parameter identification and calibration & Identify parameter values that explain the recorded observations & Transport, kinetic, geometric, or thermal parameters with specified bounds & Successful solve, exported observables, and comparison with the recorded observations \\
Forward prediction & Predict a new protocol or operating condition with a fixed model & Voltage, state of charge, capacity, or lifetime curve or map & Prediction metric on the specified evaluation protocol \\
Mechanism and multiphysics & Determine whether an added physical process is needed and correctly coupled & Executable mechanism, state, equation, or multiphysics coupling & Implementation check, solution check, and comparison with observations \\
Model diagnosis and procedure & Diagnose a failed model change and establish a reproducible correction & Failure diagnosis, corrected procedure, or verified negative result & Build, solve, and export checks with a record that links the cause to the correction \\
\bottomrule
\end{tabularx}
\end{table}

This inventory defines the evidence types used in the records below. It separates parameter, prediction, mechanism, and diagnosis evidence before the counts are interpreted.

\subsection{Run Accounting}

The collection contains 179 recorded runs grouped into 32 task families. The cross-task analysis uses 65 records from 26 families. Nine families contain a candidate fixed before evaluation, a pre-specified comparison target, a numerical comparison obtained after the candidate was fixed, and a final decision. These counts define the records used by the screening and evaluation tables below.

\begin{table}[t]
\centering
\caption{Counts for runs, task families, and evaluations.}
\label{tab:appendix-run-accounting}
\small
\setlength{\tabcolsep}{3pt}
\begin{tabularx}{\linewidth}{@{}>{\raggedright\arraybackslash}p{0.72\linewidth}>{\centering\arraybackslash}X@{}}
\toprule
Quantity & Count \\
\midrule
Recorded runs & 179 \\
Runs for application tasks & 64 \\
Runs for physics-based battery tasks & 115 \\
Task families in the recorded run collection & 32 \\
Task families in the cross-task analysis set & 26 \\
Run-level records in the cross-task analysis set & 65 \\
Task families with a complete evaluation record & 9 \\
\bottomrule
\end{tabularx}
\end{table}

The accounting table sets the scope for the analysis that follows. The 179 runs describe the full collection. The 65 records and nine complete task families contain the comparison chains.

\subsection{Cross-Task Evaluation of Scientific Decisions}
\label{sec:appendix-cross-task-validation}

This analysis follows each fixed model through a later comparison and final decision. Nine records contain the complete sequence. Four conclusions apply to the evaluated conditions. Four conclusions are narrowed to the part supported by the comparison. One proposed general mechanism is rejected. The tables below give the comparison and conclusion for each record.

\subsubsection{Task-Family Screening}

The analysis groups 65 records into 26 task families. A complete comparison record contains a stated conclusion, a fixed candidate model, a comparison target, a numerical result, and a final decision. A model file or a complete implementation with recorded parameter values identifies the candidate. The target can be a specified condition, an external dataset, another cell or time interval, or a fixed external reference. Nine task families contain a complete record. The analysis keeps one complete record from each family and checks whether its numerical result agrees with the recorded decision.

The candidate version is fixed before the comparison and remains fixed through the final decision. A comparison that changes a parameter, model structure, threshold, or candidate choice remains part of model construction. Each comparison result remains linked to the model version that produced it. In the table below, SOH means state of health. LAM-NE and LAM-PE mean loss of active material in the negative and positive electrodes. UDDS means Urban Dynamometer Driving Schedule. SEI means solid-electrolyte interphase.

\begin{table}[!htbp]
\centering
\caption{Completeness check for the 26 task families in the cross-task evaluation.}
\label{tab:appendix-family-screen}
\scriptsize
\setlength{\tabcolsep}{3.5pt}
\renewcommand{\arraystretch}{1.08}
\begin{tabularx}{\linewidth}{@{}>{\raggedright\arraybackslash}p{0.20\linewidth}>{\raggedright\arraybackslash}p{0.51\linewidth}>{\centering\arraybackslash}p{0.08\linewidth}>{\raggedright\arraybackslash}X@{}}
\toprule
Outcome & Task families & Count & Reason \\
\midrule
Complete evaluation & Molicel P42A transfer; LG M50T diagnostic-to-another-condition; degradation-mode prediction; Panasonic 18650PF drive-cycle evaluation; pouch-cell thermal prediction; LG M50-family cross-dataset transfer; COBRAPRO UDDS; CPG-SPMT FUDS; SEI mechanism competition & 9 & Fixed candidate, comparison target, numerical result, final decision, and supported conditions are available. \\
Comparison used for later model changes & Panasonic multi-temperature model; LG M50 full-matrix model & 2 & The comparison later enters model development, or the final conclusion uses a metric from an earlier model. \\
Development record & Lithium-plating temperature--rate boundary; lithium-plating risk over temperature and rate & 2 & The result contributes to threshold construction or has no matching metallic-lithium outcome. \\
No comparison result & Early-cycle lifetime surrogate; missing-data stop; model-family ablation; phase-field dead-lithium reconstruction; parameter-modification probe; NMC811 material studies; thermal smoke test; NMC811 plating studies; generic plating studies; Samsung 21700-50E calibration; SOC export; rate-capability sweep; SEI degradation demonstrations & 13 & At least one required part of a complete evaluation record is absent. \\
\midrule
Total & & 26 & \\
\bottomrule
\end{tabularx}
\end{table}

The screening table separates complete evaluations from model-development and partial records. The nine complete records support the summary above.

\subsubsection{Included Evaluation Chains}

The candidate model, comparison result, and final decision come from the original task records. The analysis uses three decision labels. \textsc{Tested only} assigns the conclusion to the specified condition. \textsc{Limited} retains the supported part of a broader prediction or mechanism conclusion. \textsc{Unsupported} records that the proposed general conclusion was not selected. A mechanism is called universal only if it passes the stated comparison at every stage and in every cell covered by the conclusion. The decision label records the scope of the conclusion, and the result column reports performance.

\begin{table}[!htbp]
\centering
\caption{Complete evaluation records from the task-family analysis.}
\label{tab:appendix-validation-chains}
\scriptsize
\setlength{\tabcolsep}{3.5pt}
\renewcommand{\arraystretch}{1.08}
\begin{tabularx}{\linewidth}{@{}>{\raggedright\arraybackslash}p{0.20\linewidth}>{\raggedright\arraybackslash}p{0.30\linewidth}>{\centering\arraybackslash}p{0.13\linewidth}>{\raggedright\arraybackslash}X@{}}
\toprule
Task & Comparison result & Decision & Conclusion supported by the result \\
\midrule
Molicel P42A 25$^\circ$C C/3 forward comparison & 44 segments; 8,827 points; 18.258\,mV RMSE & Tested only & Specified 25$^\circ$C C/3 segments \\
LG M50T prediction on another condition & 3,610 points; 44.878\,mV RMSE & Tested only & Prediction on the specified additional condition \\
LG M50T degradation-mode prediction & RMSE on another condition: SOH 5.893\%; LAM-NE 3.718\%; LAM-PE 5.212\%; resistance 4.323\,m$\Omega$ & Limited & Derived degradation quantities \\
Panasonic 18650PF drive-cycle evaluation & 35 cases; mean RMSE 71.107\,mV; 34 full windows and 1 valid early cutoff & Tested only & Specified additional drive-cycle set \\
Reduced pouch-cell thermal model & Surface-mean temperature 2.461\,K; voltage 164.593\,mV RMSE & Limited & Scalar surface-mean temperature \\
LG M50-family cross-dataset transfer & 146 points; 152.077\,mV RMSE & Limited & Source-data calibration \\
COBRAPRO LG M50T UDDS prediction & 18,835 points; 10.169\,mV RMSE & Tested only & UDDS prediction \\
CPG-SPMT A123 FUDS reproduction & 34.994\,mV mean; 0.023\,mV worse than official baseline & Limited & Source result reproduction \\
SEI mechanism competition & 0 of 3 mechanisms pass the stated cross-stage and cross-cell comparison & Unsupported & No universal cross-stage and cross-cell mechanism selected \\
\bottomrule
\end{tabularx}
\end{table}

Together, these records show how comparison results determine the conditions supported by each conclusion. The negative solid-electrolyte interphase (SEI) result is also evidence. None of the tested candidates passes the stated comparison across all stages and cells. The system therefore rejects the proposed general mechanism.

Throughout the appendix, EvoSim defines the scientific target, records a candidate modification or modification set, and tests that candidate under the relevant conditions. The result determines which model is selected and which candidate is considered next. The tables and figures report the evidence for model construction, procedure reuse, and the conditions supported by each conclusion.

\section{Reproducibility}
\label{sec:appendix-reproducibility}

\paragraph{Language-model execution.}
Each controlled comparison uses the same language-model endpoint and model version in all conditions. The three-condition experiment uses \texttt{gpt-5.5}. Its main comparison uses the first three attempts. The system prompt, task prompt, starting model, input files, tools, turn limits, attempt accounting, and model checks are fixed before execution. The later Baseline and Naive Memory attempts are reported as a separate persistence check. The provider uses the same sampling policy in every condition. Temperature and top-$p$, which control response sampling, receive the same settings. The research record is the only input that differs among the three conditions.

\paragraph{Numerical execution.}
Within each controlled comparison, the compared conditions use the same COMSOL version, solver backend, starting model, geometry, mesh, initial and boundary conditions, time grid, solver settings, output expressions, and error calculation. These quantities are stored with each candidate. The dynamic-voltage state calculation uses a relative solver tolerance of $10^{-8}$. The lithium-plating campaign uses a relative solver tolerance of $10^{-5}$. A fixed candidate can therefore be solved again without another language-model decision. The implementation check confirms the active equations and model states. The solution check follows the requested-horizon rule or the event-completion rule defined for the task before execution. It also checks finite values and the required output files.

\paragraph{Saved records.}
Each run stores the task and input-file hashes, repository revision, language-model name, runtime policy, solver budget, submitted model, exported numerical results, metric calculation, and final review decision. Local and remote solver jobs use the same directory and output structure. These records identify the model and inputs behind every reported result. Once a candidate is fixed, the saved model, inputs, and solver settings determine its numerical result. The selected candidate can therefore be solved again directly from the saved files.

\section{Additional Reusable Modeling Procedures}
\label{sec:appendix-capability-atlas}

Scientific Evolution saves a procedure when an investigation supports a specific model action. The saved record states the result that motivated the action, the expected physical response, the required checks, and the conditions under which the action was tested. A later investigation can apply the procedure only when those conditions and checks are available:

\begin{center}
\fbox{\begin{minipage}{0.91\linewidth}
\centering
\textbf{Measured or solver result}\quad $\longrightarrow$ \quad
\textbf{Physical conclusion}\quad $\longrightarrow$ \quad
\textbf{Model action}\quad $\longrightarrow$ \quad
\textbf{Checks}\quad $\longrightarrow$ \quad
\textbf{Later use}
\end{minipage}}
\end{center}

The following fields define each record. The complete record also stores the source files and model version.

\begin{center}
\begin{minipage}{\linewidth}
\centering
\textbf{Information stored for a reusable modeling procedure.}\par
\vspace{3pt}
\scriptsize
\setlength{\tabcolsep}{3pt}
\begin{tabularx}{\linewidth}{@{}>{\raggedright\arraybackslash}p{0.22\linewidth}>{\raggedright\arraybackslash}X@{}}
\toprule
Item & Required information \\
\midrule
Starting result & Measurement difference, failed candidate, conflicting result, or solver output that motivates the procedure \\
Physical conclusion & The physical or numerical cause supported by the result \\
Model action & Equation, state update, model connection, calculation, or candidate comparison \\
Expected response & Predicted direction or quantity that can be checked after the model action \\
Checks & Model identity, active equations, state values, finite outputs, and response to a specified input \\
Supported conditions & Tasks, models, and inputs for which the evidence supports the procedure \\
Later use & Later task, changed model decision, and resulting observation \\
\bottomrule
\end{tabularx}
\end{minipage}
\end{center}

The four cases below report how a result becomes a model action that can be checked again. Each case states the tested conditions. Later use requires the new task to meet those conditions and repeat the checks.

\newtcolorbox{capabilitycase}[1]{%
  enhanced jigsaw,
  breakable,
  colback=blue!2,
  colframe=blue!45!black,
  colbacktitle=blue!8,
  coltitle=black,
  title={Reusable procedure: #1},
  title after break={Reusable procedure: #1 (continued)},
  fonttitle=\bfseries,
  boxrule=0.5pt,
  arc=2pt,
  left=6pt,
  right=6pt,
  top=5pt,
  bottom=5pt,
  pad at break*=2pt,
  before skip=8pt,
  after skip=8pt
}

\begin{capabilitycase}{Discrete state update under piecewise-constant current}
\textbf{Starting result.} The continuous four-state model solves, but its surface filling fractions and voltage differ from the discrete reference calculation.\\
\textbf{Physical conclusion.} The current is constant within each one-second sample. The electrode filling states must therefore follow the exact zero-order-hold update in Equations~\ref{eq:appendix-cpg-zoh-parameters}--\ref{eq:appendix-cpg-zoh-update}.\\
\textbf{Model action.} Apply the recurrence once per sample. Convert temperature to Kelvin once. Assign the two average and two surface filling fractions to four dependent variables in the physics-based model. Calculate voltage from these states.\\
\textbf{Checks.} Check the initial states, sample-$k$ current, temperature units, all four solver states, complete finite output, and the state--voltage equation.\\
\textbf{Results.} The RMSE against the reference calculation decreases from 88.518 to 4.1136\,mV. The same update solves 24 temperature and drive-cycle conditions with a mean reference-calculation RMSE of 4.4625\,mV. Its later use in the controlled voltage task gives 22.015\,mV RMSE against measured voltage.\\
\textbf{Later use.} An initial use converts temperature twice and fails. The next attempt applies one conversion and the four-state representation, then passes the checks.\\
\textbf{Tested conditions.} One-second sampling, the stated CPG-SPMT equations, piecewise-constant current and temperature inputs, and the stated voltage equation.
\end{capabilitycase}

\begin{capabilitycase}{Distinguishing a missing process from a parameter error}
\textbf{Starting result.} A difference from measurement remains after the input data, parameter units, boundary conditions, and numerical solution have passed their checks.\\
\textbf{Physical conclusion.} A parameter error changes the magnitude of an active response. A missing process prevents the governing equations from producing that response. An inactive coupling adds a term that does not affect the solved quantity.\\
\textbf{Model action.} First confirm that the proposed source, flux, reaction, or boundary term enters the active governing equation. Then apply an input for which physics gives a clear response direction. Change a parameter only after the equation produces that response.\\
\textbf{Checks.} Inspect the active equation and the added term. Confirm the sign and units of the predicted response. Compare the complete output with the measurement using the fixed task metric.\\
\textbf{Later use.} The result determines whether the next candidate adds a physical process, repairs a coupling, or continues parameter identification.
\end{capabilitycase}

\begin{capabilitycase}{Checking an added reaction or coupling}
\textbf{Starting result.} A candidate solves, but the target voltage or temperature does not change after a reaction or coupling is added.\\
\textbf{Physical conclusion.} A successful solve does not show that the added process is active. The new term must carry a nonzero flux, current, or heat rate and must enter the equation for the reported quantity.\\
\textbf{Model action.} Evaluate the added term over the full simulation. Trace its connection to the relevant conservation equation and output expression. Apply a load under which the term should increase or decrease.\\
\textbf{Checks.} Confirm the term's units, sign, magnitude, spatial location, and time response. Confirm that the target output changes in the predicted direction.\\
\textbf{Decision.} Retain the mechanism only when the term is active and its response passes the measurement check. Otherwise, correct the connection or reject the mechanism.
\end{capabilitycase}

\begin{capabilitycase}{Comparing heat-generation and heat-transfer mechanisms}
\textbf{Starting result.} Several mechanisms can change the cell-temperature trace. Repeated changes to one candidate do not isolate their individual effects.\\
\textbf{Physics-based model.} The temperature field satisfies $\rho c_p\,\partial T/\partial t=\nabla\!\cdot(k\nabla T)+q$. Here $\rho$ is density, $c_p$ is specific heat capacity, $k$ is thermal conductivity, and $q$ is volumetric heat generation. Boundary heat transfer supplies the thermal boundary condition.\\
\textbf{Model action.} Keep the starting model, input, and temperature metric fixed. Build separate candidates for particle-size effects, electrochemical polarization heat, boundary heat flux, and convection. Compare them before combining mechanisms.\\
\textbf{Checks.} Confirm that each candidate changes the intended heat source or boundary term. Solve every candidate over the same time range. Compare every complete temperature trace with the same measurements.\\
\textbf{Later use.} EvoSim records this comparison during the thermal-mechanism investigation and applies it in the later P2D--thermal task. The number of sequential model changes decreases from 10 to 1.\\
\textbf{Tested conditions.} Tasks that can isolate candidate heat-generation or heat-transfer terms under one starting model and one temperature metric.
\end{capabilitycase}